%% file: DecoMoE.tex
\documentclass{article} 
\usepackage{iclr2027_conference,times}

\input{math_commands.tex}

\usepackage{hyperref}
\usepackage{url}

\usepackage{graphicx}

\usepackage{algorithm}
\usepackage{algorithmic}

\usepackage{booktabs}
\usepackage{multirow}

\usepackage{amsmath}
\usepackage{amssymb}

\title{
DecoMoE: Decoupling Visual Propagation and Expert Computation for
Efficient Multimodal MoE Inference
}

\author{
Xudong Tan$^{1}$\thanks{Equal contribution.}
\quad
Peng Ye$^{2}$\footnotemark[1]
\quad
Ming Xie$^{1}$\footnotemark[1]
Chenyu Huang$^{1}$
\quad
Yaoxin Yang$^{1}$
\quad
Jiayuan Fan$^{1}$
\quad
Tao Chen$^{1}$\thanks{Corresponding author: \texttt{eetchen@fudan.edu.cn}}
\\[4pt]
$^{1}$Fudan University
\qquad
$^{2}$The Chinese University of Hong Kong
}

\iclrfinalcopy

\hypersetup{
    pdftitle={
        DecoMoE: Decoupling Visual Propagation and Expert Computation
        for Efficient Multimodal MoE Inference
    },
    pdfauthor={
        Xudong Tan, Peng Ye, Ming Xie, Chenyu Huang,
        Yaoxin Yang, Jiayuan Fan, Tao Chen
    }
}

\begin{document}

\maketitle


\fancyhead{}
\renewcommand{\headrulewidth}{0pt}


\input{sec/0_abstract}

\input{sec/1_introduction}

\input{sec/2_Related_Work}
\input{sec/3_Methodology}

\input{sec/4_Experiments}

\input{sec/5_Conclusion}









\bibliography{iclr2027_conference}
\bibliographystyle{iclr2027_conference}


\appendix
\input{sec/Appendix}

\end{document}

%% file: math_commands.tex
\usepackage{amsmath,amsfonts,bm}

\def\eqref#1{equation~\ref{#1}}

\def\1{\bm{1}}

\DeclareMathAlphabet{\mathsfit}{\encodingdefault}{\sfdefault}{m}{sl}
\SetMathAlphabet{\mathsfit}{bold}{\encodingdefault}{\sfdefault}{bx}{n}



%% file: sec/0_abstract.tex
\begin{abstract}
Multimodal mixture-of-experts (MoE) models combine sparse expert activation with visual-language capabilities, yet their inference remains costly because long visual-token sequences repeatedly incur attention, routing, dispatch, and expert-MLP computation. Existing methods typically compress either the token or expert dimension, leaving redundancy along the other. Our analysis reveals two complementary regularities: the depth required for visual propagation varies across inputs, while text-token routing exhibits concentrated and recurrent expert-importance patterns. Based on these observations, we propose DecoMoE, a two-dimensional structured compression framework that decouples visual propagation from expert computation. The Sample-Adaptive Visual Boundary (SAVB) predicts an input-dependent visual-exit layer at which the visual-token block is removed. The Routing-Calibrated Expert Prefix (RCEP) reorders experts offline using text-token routed mass and, from this predicted exit layer onward, retains at each MoE layer the shortest contiguous prefix covering a target routed-mass fraction. We evaluate DecoMoE on Qwen3-VL-MoE and InternVL3.5-30B-A3B across six benchmarks. On Qwen3-VL-MoE, DecoMoE retains 97.91\% of dense-baseline performance while reducing computation from 27.06 to 16.73 TFLOPs and latency from 0.44 to 0.26 seconds, yielding a 1.69× speedup. Code will be available at https://github.com/ShawnTan86/DecoMoE.
\end{abstract}

%% file: sec/1_introduction.tex
\section{Introduction}
\label{sec:introduction}

Multimodal large language models (MLLMs) extend language models to visual understanding and reasoning~\cite{liu2023visual,wang2024qwen2vl,chen2024internvl}, while mixture-of-experts (MoE) architectures scale model capacity through sparse token-wise expert activation~\cite{jiang2024mixtral,wu2024deepseekvl2}. However, sparse activation alone does not make multimodal MoE inference inexpensive: long visual-token sequences may propagate through many decoder layers, repeatedly incurring attention, routing, dispatch, and expert-MLP computation. Multimodal MoE inference therefore presents a two-dimensional compression problem: the token dimension determines the sequence length and its associated attention cost, whereas the expert dimension determines the expert span available to routing and the corresponding dispatch and MLP costs. Compressing only one dimension leaves substantial redundancy along the other.

\begin{figure*}[t]
    \centering

    \begin{minipage}[t]{0.49\textwidth}
        \centering
        \includegraphics[width=\linewidth]{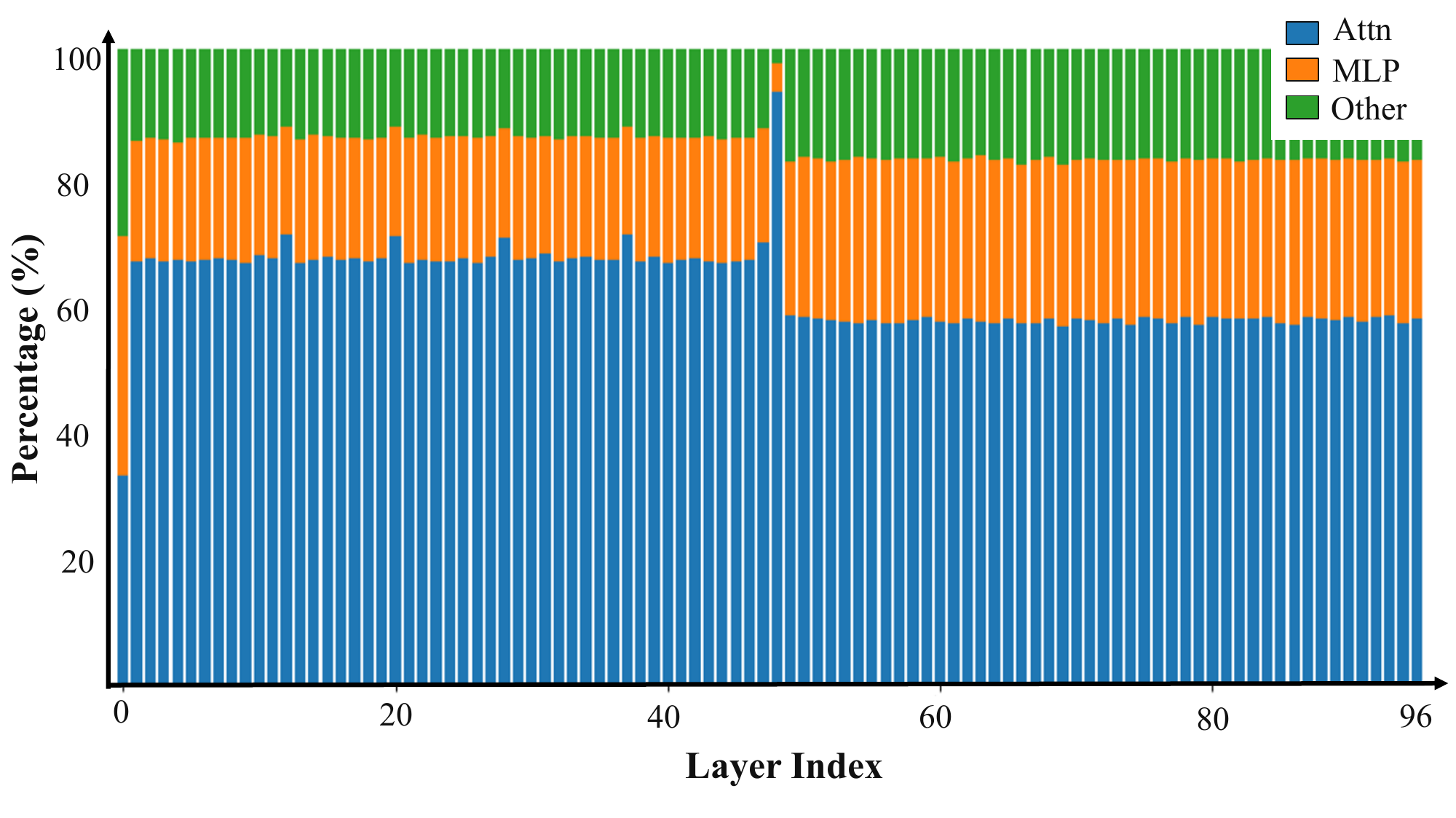}
        \vspace{2pt}
        
        \textbf{(a) Per-layer latency composition}
    \end{minipage}
    \hfill
    \begin{minipage}[t]{0.49\textwidth}
        \centering
        \includegraphics[width=\linewidth]{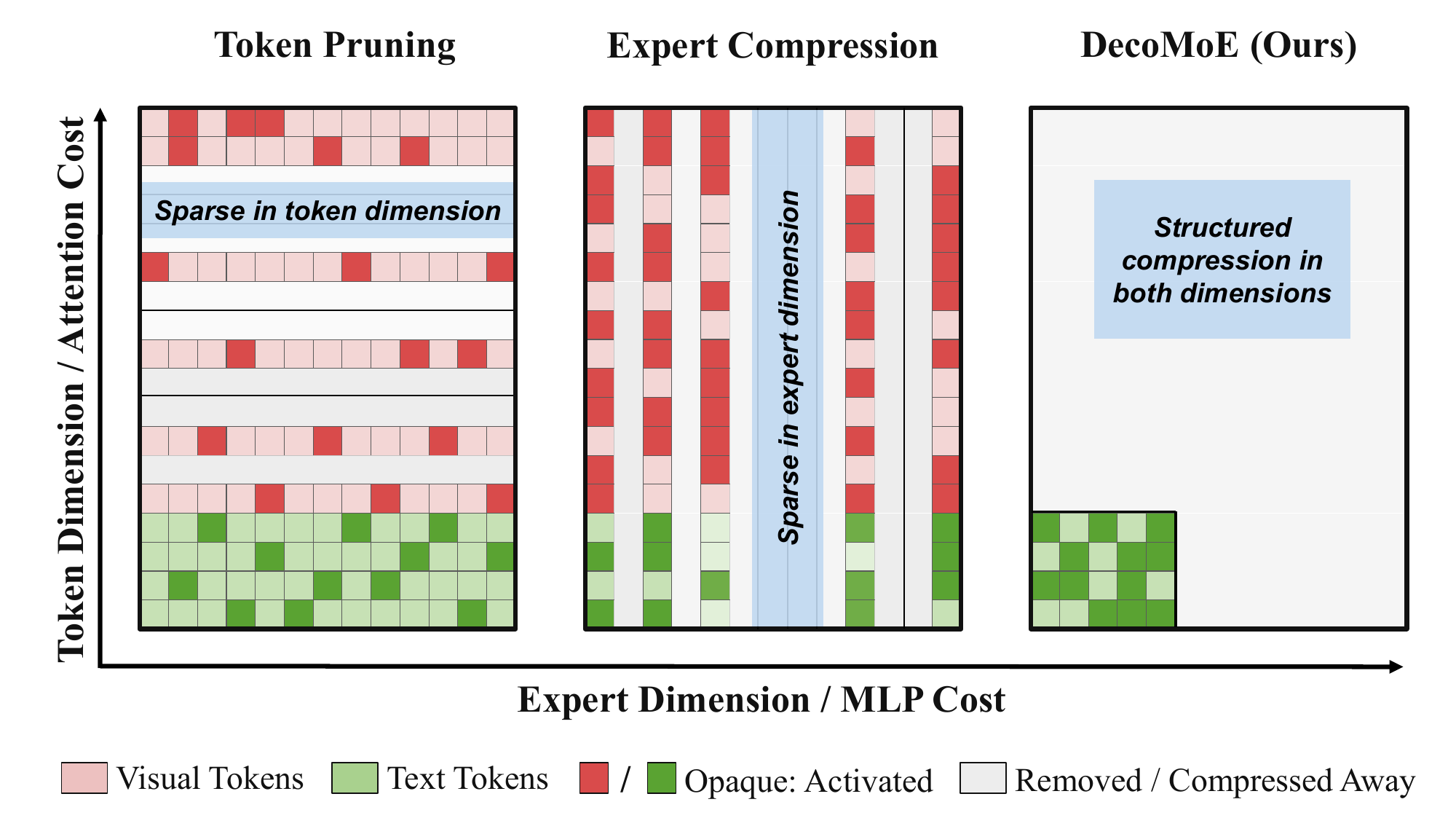}
        \vspace{2pt}
        
        \textbf{(b) Compression across token and expert dimensions}
    \end{minipage}

    \caption{
    Computation characteristics and structured compression of multimodal MoE inference.
    (a) Per-layer latency composition across one prefill pass and one decoding pass, shown as two consecutive 48-layer segments. Except for a small number of boundary-layer outliers, Attention and the expert MLP jointly account for more than 80\% of the measured inference latency.
    (b) Comparison of compression across the token and expert dimensions, corresponding to attention and expert-MLP costs. Token pruning and expert compression sparsify one dimension, whereas DecoMoE removes the visual-token block and retains a contiguous prefix of experts. Opaque cells denote routed activations; gray regions denote removed computation.
    }
    \label{fig:method_and_latency}
\end{figure*}

Figure~\ref{fig:method_and_latency}(b) places existing approaches in this two-dimensional computation space. Visual-token pruning methods such as FastV and SparseVLM sparsify the token dimension within the language model~\cite{chen2024fastv,zhang2025sparsevlm}, while whole-block methods such as VTW and DyVTE remove the complete visual-token block at a selected layer~\cite{lin2025vtw,wu2026accelerating}. Both reduce attention-side computation but retain the original expert span. Conversely, expert pruning and skipping methods reduce the expert footprint or activated expert computation~\cite{lu2024experts,huang2025shapleymoe}, while typically preserving the full multimodal sequence. Recent multimodal MoE methods combine token reduction with routing-aware expert activation or modality-specific expert skipping~\cite{xia2025fastmmoe,huang2026modes}, suggesting that both dimensions contain reducible redundancy. Yet reducing both dimensions is not sufficient by itself: the token-side decision should reflect when visual propagation can stop for the current input, while expert-side compression should organize routing decisions into a structure that supports efficient execution. Existing approaches do not explicitly use a sample-dependent visual boundary to initiate text-calibrated, physically contiguous expert-prefix execution.

Our diagnostics reveal two complementary regularities (Figure~\ref{fig:decoupling-evidence}). First, the earliest layer at which the complete visual-token block can be removed while preserving the answer varies across inputs, and later exits typically remain in a stable low-perplexity region. This motivates controlling visual propagation at the sample level rather than with one global layer. Second, text-token routing exhibits higher routed-mass concentration and larger cross-case expert overlap than visual-token routing. This suggests that expert importance for the subsequent text-only computation should be calibrated from text-aligned routing statistics rather than from mixed multimodal routing.

Motivated by these findings, we propose \textbf{DecoMoE}, a routing-aware framework that decouples the sample-dependent depth of visual propagation from the recurring expert structure used for subsequent text-only computation. DecoMoE contains two complementary components. The \textbf{Sample-Adaptive Visual Boundary (SAVB)} evaluates a lightweight gate from the hidden states entering sparse candidate layers. If the gate triggers at layer $l$, SAVB removes the complete visual-token block before self-attention in that layer. The gate is supervised by forced-exit correctness and first-token negative log-likelihood while the multimodal backbone remains frozen. Starting from the MoE block of the same layer, the \textbf{Routing-Calibrated Expert Prefix (RCEP)} follows a layer-wise Mean-Mass order calibrated offline from text-token routed mass, physically reorders the corresponding router outputs and expert parameters, and dynamically retains the shortest prefix covering a target fraction of the current activated routed mass. Native token-wise Top-$k$ routing and expert-MLP execution are then restricted to the prefix selected independently at layer $l$ and each subsequent layer.

We evaluate DecoMoE on Qwen3-VL-MoE and InternVL3.5-30B-A3B across six benchmarks. DecoMoE retains 97.91\% and 97.87\% of baseline performance while reducing computation from 27.06 to 16.73 and 9.86 to 5.97 TFLOPs, respectively, with measured speedups of 1.69$\times$ and 1.41$\times$. Additional analyses examine both decoupling mechanisms: SAVB identifies sample-dependent visual boundaries after which PPL remains comparable to full visual propagation, while the Expert-Importance Decoupling Analysis shows that RCEP better identifies experts important for text-token computation, supporting its use for offline expert reordering. Our contributions are summarized as follows:
\begin{itemize}
    \item We identify two complementary regularities in multimodal MoE inference: an input-dependent visual-propagation boundary and a more concentrated, cross-case recurrent expert structure for text-token routing. These observations motivate separate treatment of visual propagation and expert importance.
    \item We introduce DecoMoE, a two-dimensional structured compression framework that decouples visual propagation from expert computation, using a sample-adaptive boundary for contiguous visual-token removal and text-routing calibration to organize experts by importance into a contiguous prefix for structured expert compression.
    \item We evaluate DecoMoE on Qwen3-VL-MoE and InternVL3.5-30B-A3B across six benchmarks. On Qwen3-VL-MoE, DecoMoE achieves a 1.69$\times$ measured speedup with a 2.09\% average performance reduction relative to the dense baseline.
\end{itemize}

%% file: sec/2_Related_Work.tex
\section{Related Work}
\label{sec:related-work}

\subsection{Multimodal Mixture-of-Experts Models}

MLLMs align visual encoders with causal language models through projectors and visual instruction tuning~\cite{liu2023visual,bai2023qwenvl,wang2024qwen2vl}. In parallel, sparse MoE models such as Mixtral and DeepSeekMoE expand model capacity through token-wise conditional routing~\cite{jiang2024mixtral,dai2024deepseekmoe}. This architecture has been extended to multimodal reasoning by MoE-LLaVA, DeepSeek-VL2, Kimi-VL, Qwen3-VL, and InternVL3.5~\cite{lin2024moellava,wu2024deepseekvl2,kimiteam2025kimivl,bai2025qwen3vl,wang2025internvl35}. Compared with dense MLLMs, their inference cost is jointly determined by sequence length and routed expert computation, while visual and textual tokens may induce different expert loads. Although sparse routing limits the experts activated by each token, long visual sequences still traverse attention, routing, dispatch, and expert MLPs at every retained layer.

\subsection{Efficient Multimodal Mixture-of-Experts Models}

Most MLLM acceleration methods reduce the visual-token sequence. Front-end approaches compress visual features before language decoding~\cite{li2024tokenpacker,shang2025llavaprumerge,yang2025visionzip}, whereas decoder-side methods prune or merge tokens using attention, statistical, visual, optimization, or information-preservation criteria~\cite{chen2024fastv,ye2024fitprune,zhang2025sparsevlm,xing2025pyramiddrop,zhang2025vispruner,yang2025topv,tan2025tokencarve}. Dynamic-LLaVA, LLaVA-Mini, and MustDrop learn or schedule compression across inference stages~\cite{huang2025dynamicllava,zhang2025llavamini,liu2024mustdrop}. Closest to whole-block removal, VTW withdraws visual tokens at a calibrated layer, while DyVTE predicts an input-dependent exit depth~\cite{lin2025vtw,wu2026accelerating}. These studies show that visual redundancy varies across processing stages, but their compression decisions are independent of MoE routing behavior. They therefore primarily target dense backbones or leave original expert pool unchanged after token reduction.

A complementary line compresses MoE experts through pruning, decomposition, quantization, or contribution estimation~\cite{lu2024experts,yang2024moei2,xie2024moepruner,huang2025mcmoe,huang2025shapleymoe}. More recent methods directly address multimodal MoE inference: FastMMoE combines visual-token pruning with reduced expert activation, MoDES performs layer- and modality-aware expert skipping, and MACS mitigates modality-induced load imbalance under expert parallelism~\cite{xia2025fastmmoe,huang2026modes,li2026macs}. Unlike permanent expert compression, these approaches adapt the executed computation or capacity while retaining the pretrained model structure. Routing analyses further reveal distinct visual and textual expert behaviors~\cite{xu2026seeing}. DecoMoE differs by using an input-dependent visual-propagation boundary to determine when computation becomes text-dominated, and cross-case text-routing regularity to determine which physically contiguous expert prefix is retained at each subsequent MoE layer. The two decisions are combined during inference but treated separately in their formulation and calibration.

%% file: sec/3_Methodology.tex
\begin{figure*}[t]
    \centering
    \includegraphics[width=\textwidth]{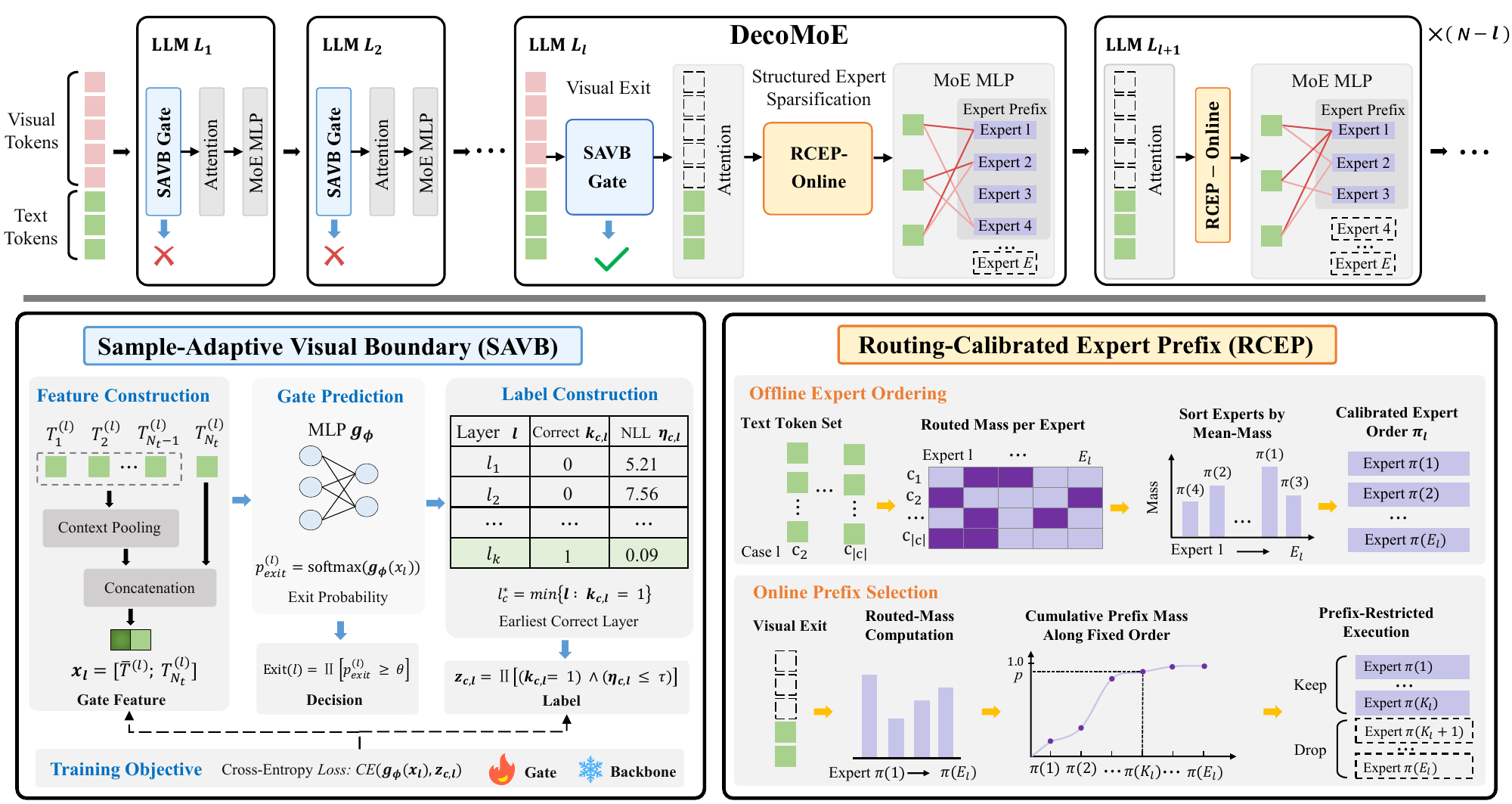}
    \vspace{-20pt}
    \caption{Overview of DecoMoE. \textbf{Top:} At each candidate-layer entrance, SAVB determines whether to remove the complete visual-token block before self-attention. Once triggered at layer $l$, RCEP is activated in the MoE block of the same layer and independently restricts the triggering layer and each subsequent MoE layer to a mass-sufficient contiguous prefix of offline-reordered experts. \textbf{Bottom left:} SAVB constructs its gate feature from the pooled preceding text states and the final text state, and is trained with labels derived from forced-exit correctness and first-token NLL while the backbone remains frozen. \textbf{Bottom right:} Offline, RCEP derives a layer-wise Mean-Mass expert order from cross-case text-token routing. Online, it aggregates the activated routed mass over tokens in each MoE invocation and retains the shortest prefix covering the target mass fraction $p$.}
    \label{fig:method-overview}
    \vspace{-10pt}
\end{figure*}

\section{Methodology}
\label{sec:methodology}

DecoMoE is a routing-aware, two-dimensional structured compression framework for multimodal MoE inference. It decouples the boundary of visual propagation from the text-aligned expert structure used thereafter. As shown in Figure~\ref{fig:method-overview}, SAVB determines when the sequence transitions to text-only computation, while RCEP selects a contiguous prefix of reordered experts at each MoE layer thereafter. The visual boundary and layer-wise prefix lengths remain input-dependent, while their execution is structurally regular along the token and expert dimensions.

\subsection{Principle}
\label{subsec:decoupling-principle}

Consider a multimodal MoE decoder with $N$ layers, indexed by $l\in\{1,\ldots,N\}$. Let the multimodal input sequence be $X=[X_v;X_t]$, where $X_v$ is a contiguous visual-token block and $X_t$ is the text-token sequence. At the entrance to decoder layer $l$, the hidden sequence is $H_l=[V^{(l)};T^{(l)}]$. Let $\mathcal{E}_l$ denote its expert pool and $E_l=|\mathcal{E}_l|$ the number of experts. For token position $i$, the MoE router produces logits $\mathbf{a}_{l,i}$ and assigns probability
\begin{equation}
    \rho_{l,i,e}
    =
    \frac{\exp(a_{l,i,e})}
    {\sum_{e'\in\mathcal{E}_l}\exp(a_{l,i,e'})},
    \qquad e\in\mathcal{E}_l .
    \label{eq:decoupling-router-prob}
\end{equation}
These quantities characterize two coupled dimensions of MoE computation: visual-token propagation determines how long $V^{(l)}$ remains in the sequence, while $\boldsymbol{\rho}_{l,i}$ determines how computation is allocated across experts.

\begin{figure*}[!t]
    \centering
    \includegraphics[width=\textwidth]{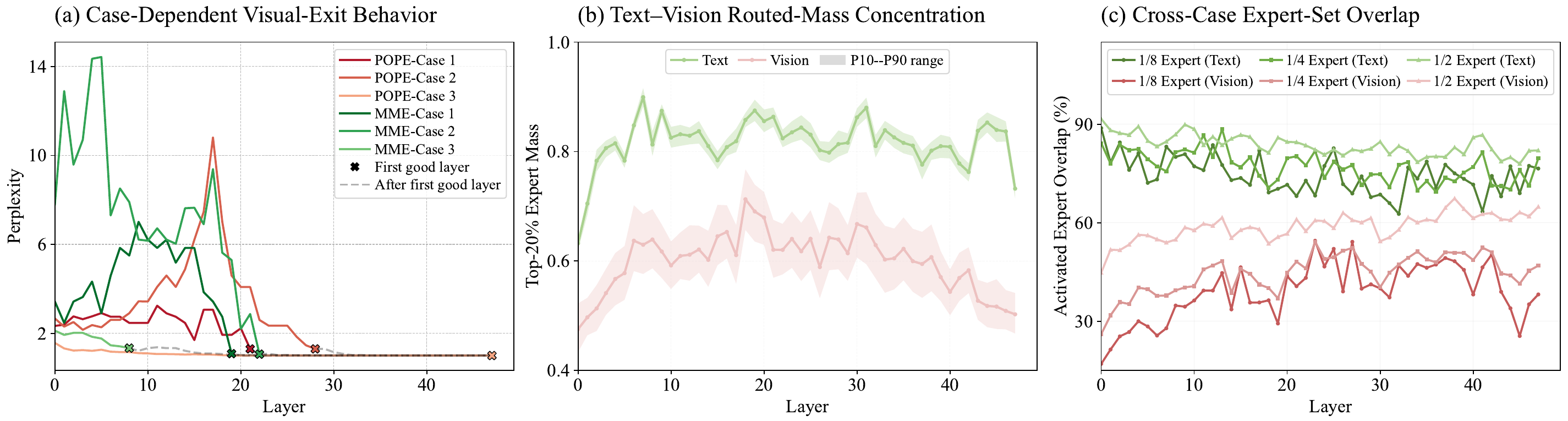}
    \vspace{-20pt}
    \caption{Behavioral diagnostics for decoupled compression.
    (a) Forced visual-exit perplexity trajectories for different cases. Each curve reports the perplexity of the reference answer when visual tokens are removed after different prefill layers, and each marker indicates the earliest prefill layer at which forced visual exit still produces the correct answer.
    (b) Dataset-level fractions of routed mass covered by the top 20\% of experts for visual and text tokens across layers.
    (c) Cross-case proportions of repeated experts for visual and text tokens across layers under fixed expert-retention ratios of $1/8$, $1/4$, and $1/2$.}
    \label{fig:decoupling-evidence}
    \vspace{-10pt}
\end{figure*}

Figure~\ref{fig:decoupling-evidence} motivates decoupling the two dimensions. Figure~\ref{fig:decoupling-evidence}(a) shows that the earliest prefill layer preserving answer correctness after visual-token removal varies across cases, while perplexity (PPL) often decreases after this first-good layer and remains below that of earlier exits. Figures~\ref{fig:decoupling-evidence}(b) and~\ref{fig:decoupling-evidence}(c) show that text-token routing exhibits higher routed-mass concentration and larger cross-case proportions of repeated experts than visual-token routing, suggesting more recurrent expert-importance patterns for text tokens. These observations motivate visual-propagation decoupling, where SAVB estimates an input-dependent visual-exit boundary rather than using a fixed propagation depth, and text-aligned expert-importance decoupling, where RCEP estimates expert importance from cross-case text-token routing statistics, constructs a layer-wise expert order, and selects an input-dependent expert prefix during online inference.

\subsection{Sample-Adaptive Visual-Propagation Decoupling}
\label{subsec:visual-propagation-decoupling}

The Sample-Adaptive Visual Boundary (SAVB) module predicts an input-dependent visual-exit layer under a calibrated confidence criterion. To limit overhead, SAVB is evaluated only at a sparse candidate set $\mathcal{G}\subseteq\{1,\ldots,N\}$, with candidates placed at intervals of four decoder layers.

At candidate layer $l\in\mathcal{G}$, let $T_j^{(l)}\in\mathbb{R}^{d}$ denote the incoming hidden state of the $j$-th text token, where $j\in\{1,\ldots,n\}$. Following the feature form used by DyVTE~\cite{wu2026accelerating}, SAVB combines the final text state with the pooled preceding context:
\begin{equation}
    \bar{T}^{(l)}
    =
    \frac{1}{n-1}\sum_{j=1}^{n-1}T_j^{(l)},
    \qquad
    x_l=[\bar{T}^{(l)};T_n^{(l)}].
    \label{eq:gate-feature}
\end{equation}
Here, $x_l\in\mathbb{R}^{2d}$ concatenates the pooled context with the final text state. A two-class MLP $g_\phi$ produces the visual-exit probability and decision:
\begin{equation}
    p_{\mathrm{exit}}^{(l)}
    =
    \operatorname{softmax}(g_\phi(x_l))_1,
    \qquad
    \operatorname{Exit}(l)
    =
    \mathbb{I}[p_{\mathrm{exit}}^{(l)}\ge\theta].
    \label{eq:exit-decision}
\end{equation}
Here, $\theta$ is the online gate threshold. If SAVB triggers at the entrance to layer $l$, the complete visual-token block is removed before self-attention in that layer. Consequently, both attention and MoE computation in layer $l$ operate on the shortened text-only sequence. RCEP is enabled for the MoE block of layer $l$ and remains active in every subsequent MoE layer $m\in\{l,\ldots,N\}$.

Gate supervision is constructed by first generating a no-exit reference answer and then forcing visual exit at every candidate layer. For calibration case $c$, let $\kappa_{c,l}\in\{0,1\}$ indicate whether the forced-exit output is correct under the dataset-specific evaluation rule, and let $\eta_{c,l}$ denote the negative log-likelihood (NLL) of the reference answer's first token. The earliest correct candidate layer is
\begin{equation}
    l_c^\star
    =
    \min\{l\in\mathcal{G}:\kappa_{c,l}=1\}.
    \label{eq:first-good-layer}
\end{equation}
The corresponding NLL values define a percentile-based confidence cutoff:
\begin{equation}
    \tau(P_{\mathrm{NLL}})
    =
    \operatorname{Quantile}_{P_{\mathrm{NLL}}}
    \left(
    \{\eta_{c,l_c^\star}\mid l_c^\star\ \text{exists}\}
    \right),
    \label{eq:nll-percentile}
\end{equation}
where $P_{\mathrm{NLL}}\in(0,1)$ is the selected percentile level. Each case--layer pair is then labeled independently:
\begin{equation}
    z_{c,l}
    =
    \mathbb{I}\!\left[
        \kappa_{c,l}=1
        \ \wedge\
        \eta_{c,l}\le\tau(P_{\mathrm{NLL}})
    \right],
    \qquad l\in\mathcal{G}.
    \label{eq:gate-label}
\end{equation}
The first-good layer is used only to calibrate the cutoff and does not impose a monotonic labeling rule on later layers. SAVB is trained with cross-entropy on $(x_{c,l},z_{c,l})$ while the backbone multimodal MoE remains frozen.

\subsection{Text-Aligned Expert-Importance Decoupling with Routing-Calibrated Prefixes}
\label{subsec:text-aligned-expert-decoupling}

The Routing-Calibrated Expert Prefix (RCEP) module operates in the MoE block of the layer where SAVB triggers and in all subsequent MoE layers. It derives a layer-wise expert order from text-token routing statistics and materializes the calibrated order as a contiguous expert prefix for online inference. The order is fixed after offline calibration, whereas the retained prefix length is selected dynamically for every layer invocation from the visual-exit layer onward.

Using the router probabilities defined in Eq.~\eqref{eq:decoupling-router-prob}, let
\begin{equation}
    S_{c,l,i}
    =
    \operatorname{TopK}
    (\boldsymbol{\rho}_{c,l,i},k_{\mathrm{moe}})
    \label{eq:topk-expert-set}
\end{equation}
denote the $k_{\mathrm{moe}}$ experts selected for text token $i$ of calibration case $c$ at layer $l$. For the text-token set $\mathcal{T}_{c,l}$, the routed mass assigned to expert $e$ is
\begin{equation}
    M_{c,l,e}(\mathcal{T}_{c,l})
    =
    \frac{1}{|\mathcal{T}_{c,l}|}
    \sum_{i\in\mathcal{T}_{c,l}}
    \frac{
        \mathbb{I}[e\in S_{c,l,i}]
        \rho_{c,l,i,e}
    }{
        \sum_{e'\in S_{c,l,i}}
        \rho_{c,l,i,e'}
    }.
    \label{eq:routed-mass}
\end{equation}

Let $\mathcal{C}$ denote the calibration cases used for offline expert ordering. Figure~\ref{fig:decoupling-evidence}(c) shows substantial overlap among text-routed experts, suggesting a recurrent expert-importance structure across cases. We therefore calibrate a shared layer-wise expert order from mean routed mass:
\begin{equation}
    \pi_l
    =
    \operatorname{argsort}_{e\in\mathcal{E}_l}
    \left(
        \frac{1}{|\mathcal{C}|}
        \sum_{c\in\mathcal{C}}
        M_{c,l,e}(\mathcal{T}_{c,l});
        \mathrm{desc}
    \right).
    \label{eq:mean-mass-order}
\end{equation}

After calibration, the permutation $\pi_l$ is applied consistently to the router output dimensions and corresponding expert parameters. The reordered parameters are loaded before inference, placing experts with larger mean text-token routed mass at earlier physical positions so that each leading segment forms a contiguous prefix. This changes only the parameter order, not the learned values.

During inference, let $m\in\{l,\ldots,N\}$ denote the current MoE layer after SAVB triggers at layer $l$. RCEP determines its prefix length from the aggregate activated mass of tokens in the current invocation. Let $q_{m,i,e}$ denote the softmax probability over all $E_m$ reordered experts, and define
\begin{equation}
    R_{m,i}(x)
    =
    \operatorname{TopK}
    (\mathbf{q}_{m,i},k_{\mathrm{moe}}).
    \label{eq:runtime-topk-set}
\end{equation}
The activated mass assigned to expert $e$ is
\begin{equation}
    s_{m,e}(x)
    =
    \sum_{i\in\mathcal{I}_m(x)}
    \frac{
        \mathbb{I}[e\in R_{m,i}(x)]
        q_{m,i,e}
    }{
        \sum_{e'\in R_{m,i}(x)}
        q_{m,i,e'}
    },
    \label{eq:runtime-activated-mass}
\end{equation}
where $\mathcal{I}_m(x)$ denotes the token positions processed in this invocation. Since experts are reordered offline, no input-specific reordering occurs online.

Given a target routed-mass fraction $p\in(0,1]$, the prefix length at layer $m$ is
\begin{equation}
\begin{aligned}
    \widehat{K}_m(x;p)
    &=
    \min\left\{
        K\in\{1,\ldots,E_m\}
        \;\middle|\;
        \frac{\sum_{e=1}^{K}s_{m,e}(x)}
             {\sum_{e=1}^{E_m}s_{m,e}(x)}
        \ge p
    \right\},\\
    K_m(x;p)
    &=
    \max\{k_{\mathrm{moe}},\widehat{K}_m(x;p)\}.
\end{aligned}
    \label{eq:runtime-prefix}
\end{equation}
Thus, $p$ controls the aggregate activated mass covered by the prefix, and $K_m$ may vary across inputs, layers, and invocations. Once determined, RCEP restricts the reordered router logits to the first $K_m$ experts, recomputes softmax and Top-$k_{\mathrm{moe}}$ selection within the prefix, and renormalizes the selected weights. Since $K_m \geq k_{\mathrm{moe}}$, each token remains dispatched to $k_{\mathrm{moe}}$ experts, with experts outside the prefix excluded from routing and MLP evaluation. Router logits are computed over all $E_m$ reordered experts to estimate $K_m$; final routing uses only the first $K_m$ entries.

%% file: sec/4_Experiments.tex
\begin{table*}[!t]
\centering
\small
\setlength{\tabcolsep}{4pt}
\renewcommand{\arraystretch}{1.25}
\resizebox{\textwidth}{!}{%
\begin{tabular}{c|cccccc|ccc}
\toprule
Method
& MMBench
& ScienceQA
& AI2D
& AOKVQA
& MMMU
& HallusionBench
& Avg. Ret.
& Avg. TFLOPs
& Latency (s) \\
\midrule

\multicolumn{10}{c}{\textbf{Qwen3-VL-30B-A3B}} \\
\midrule
Baseline
& 88.98
& 90.90
& 87.21
& 87.60
& 58.26
& 74.97
& 100.00\%
& 27.06
& 0.44 (1.00$\times$) \\

FastV
& \underline{87.74}
& 87.31
& 82.93
& 86.11
& \textbf{57.21}
& 71.67
& \underline{96.32\%}
& 16.65
& 0.36 (0.82$\times$) \\

SparseVLM
& 87.59
& \underline{88.51}
& \underline{83.78}
& 86.29
& \underline{56.35}
& 69.63
& 96.02\%
& 16.21
& 0.37 (0.84$\times$) \\

DyVTE
& 60.97
& 87.97
& 82.74
& \underline{87.07}
& 52.53
& \textbf{74.21}
& 91.63\%
& 18.29
& 0.37 (0.84$\times$) \\

FastMMoE
& 82.18
& 85.33
& 74.61
& 80.17
& 54.35
& 67.09
& 90.35\%
& 18.88
& 1.56 (3.56$\times$) \\

DecoMoE
& \textbf{87.75}
& \textbf{90.80}
& \textbf{85.62}
& \textbf{87.51}
& 56.16
& \underline{71.92}
& \textbf{97.91\%}
& 16.73
& 0.26 (0.59$\times$) \\

\midrule
\multicolumn{10}{c}{\textbf{InternVL3.5-30B-A3B}} \\
\midrule
Baseline
& 88.85
& 94.08
& 86.79
& 87.95
& 59.60
& 69.70
& 100.00\%
& 9.86
& 1.50 (1.00$\times$) \\

FastV
& 87.55
& \textbf{93.28}
& 84.53
& 86.70
& \textbf{58.02}
& \textbf{66.98}
& 97.32\%
& 5.86
& 1.77 (1.18$\times$) \\

SparseVLM
& \underline{87.70}
& \underline{93.19}
& \underline{84.83}
& \underline{87.31}
& 57.24
& 65.04
& \underline{97.44\%}
& 5.03
& 1.85 (1.23$\times$) \\

FastMMoE
& 85.67
& 92.27
& 78.28
& 84.86
& 57.23
& 63.42
& 95.05\%
& 8.02
& 1.46 (0.97$\times$) \\

DecoMoE
& \textbf{88.19}
& \textbf{93.28}
& \textbf{86.59}
& \textbf{87.94}
& \underline{57.78}
& \underline{65.51}
& \textbf{97.87\%}
& 5.97
& 1.06 (0.71$\times$) \\

\bottomrule
\end{tabular}%
}
\vspace{-5pt}
\caption{Accuracy--efficiency comparison on Qwen3-VL-MoE and InternVL3.5-30B-A3B. Avg. Ret. is the mean performance retention across six benchmarks; Avg. TFLOPs is the theoretical computation per profiled case; Latency is mean profiler CUDA time. Parentheses report latency relative to the corresponding baseline.}
\label{tab:qwen_selected_method_efficiency}
\vspace{-10pt}
\end{table*}

\section{Experiments}
\label{sec:experiments}

\subsection{Experimental Setup}
\label{sec:exp_setup}

We evaluate DecoMoE on Qwen3-VL-30B-A3B-Instruct (Qwen3-VL-MoE) and InternVL3.5-30B-A3B across MMBench, ScienceQA, AI2D, AOKVQA, MMMU, and HallusionBench, using their accuracy-like metrics. We compare with the dense baseline, FastV, SparseVLM, DyVTE, and FastMMoE under the same generation and scoring protocol within each backbone. Unless stated otherwise, SAVB evaluates hidden states every four decoder layers and is supervised on a random 50\% subset of POPE and MME, while RCEP uses $p=0.70$ and calibrates its layer-wise expert order on a random 10\% subset of POPE and MME. POPE and MME are therefore excluded from the main evaluation to avoid overlap between calibration data and reported evaluation benchmarks. The backbone remains frozen throughout, with construction details provided in the appendix. All experiments use NVIDIA H200 GPUs. Latency is measured as CUDA time with \texttt{torch.profiler}, while FLOPs are theoretical estimates detailed in the appendix. Avg. Ret. is the unweighted mean of per-dataset retention relative to the corresponding dense baseline across the six evaluation benchmarks.

\begin{figure*}[t]
    \centering
    \includegraphics[width=\textwidth]{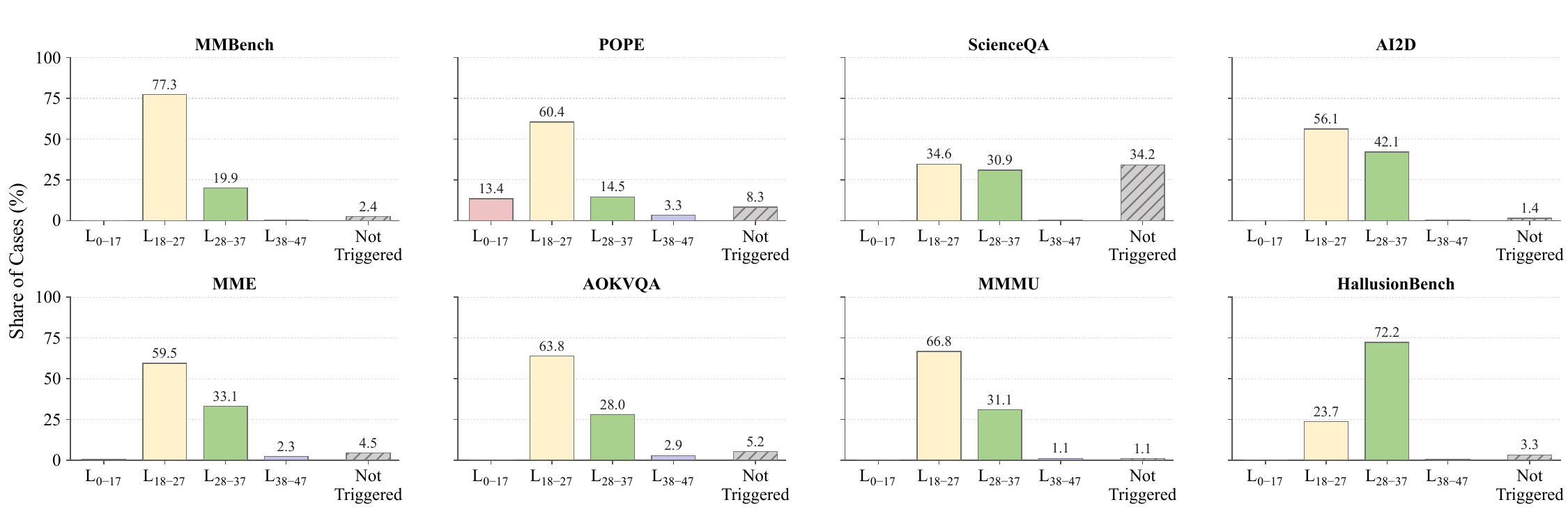}
    \vspace{-20pt}
    \caption{Dataset-wise SAVB exit distributions on Qwen3-VL-MoE. Triggered cases are grouped by zero-based layer intervals; Cases labeled Not Triggered retain visual tokens through all layers.}
    \label{fig:savb_exit_distribution}
    \vspace{-10pt}
\end{figure*}

\subsection{Main Results}
\label{sec:exp_main}

On Qwen3-VL-MoE, DecoMoE retains 97.91\% of baseline while reducing computation from 27.06 to 16.73 TFLOPs and latency from 0.44 to 0.26 seconds, corresponding to a 1.69$\times$ speedup. At similar FLOPs, its retention is 1.59 and 1.89 percentage points higher than FastV and SparseVLM, respectively. This trend also holds on InternVL3.5-30B-A3B, where DecoMoE retains 97.87\%, reduces computation from 9.86 to 5.97 TFLOPs, and lowers latency from 1.50 to 1.06 seconds (1.41$\times$). FastV and SparseVLM attain lower TFLOPs but exceed the dense latency, indicating that lower FLOPs do not necessarily translate into lower measured latency on this backbone.

\begin{table*}[!t]
\centering
\small
\setlength{\tabcolsep}{5pt}
\renewcommand{\arraystretch}{1.25}
\resizebox{\textwidth}{!}{%
\begin{tabular}{c|ccc|ccccc}
\toprule
Method
& POPE
& HallusionBench
& AI2D
& Avg. Ret.
& Avg. Exit Layer
& Avg. Prefix Size
& Avg. TFLOPs
& Latency (s) \\
\midrule
Baseline
& 89.82
& 74.97
& 87.21
& 100.00\%
& --
& 128
& 27.04
& 0.37 (1.00$\times$) \\

RCEP only
& 88.52
& 69.51
& 84.84
& 96.18\%
& 28.00
& 34.86
& 16.69
& 0.28 (0.76$\times$) \\

SAVB + Fixed Prefix
& 89.08
& 71.40
& 85.49
& 97.48\%
& 28.07
& 50
& 17.17
& 0.29 (0.79$\times$) \\

SAVB only
& 89.06
& 70.95
& 85.85
& 97.41\%
& 28.08
& 128
& 17.85
& 0.30 (0.81$\times$) \\

Full DecoMoE
& 89.27
& 71.92
& 85.62
& 97.83\%
& 28.07
& 35.01
& 17.07
& 0.26 (0.69$\times$) \\
\bottomrule
\end{tabular}%
}
\caption{Component ablation on Qwen3-VL-MoE over POPE, HallusionBench, and AI2D. RCEP only uses a fixed visual exit at layer $L_{28}$; SAVB + Fixed Prefix retains 50 experts after exit; SAVB only keeps all 128 experts.}
\label{tab:qwen_component_ablation}
\end{table*}

\subsection{Component-wise Ablation Study}
\label{sec:exp_component_ablation}

Table~\ref{tab:qwen_component_ablation} isolates visual-boundary and expert-prefix designs on POPE, HallusionBench, and AI2D. Replacing SAVB with a fixed exit at $L_{28}$ yields 96.18\% retention, whereas SAVB reaches 97.41\% while reducing computation from 27.04 to 17.85 TFLOPs. This indicates that visual-propagation depth is not well represented by a global exit layer. With SAVB fixed, the dynamic RCEP prefix averages 35.01 experts and reaches 97.83\% retention, compared with 97.48\% for a fixed 50-expert prefix. Full DecoMoE also achieves the lowest latency (0.69$\times$ baseline), showing that sample-adaptive visual exit and mass-adaptive expert selection contribute complementary reductions.

\subsection{Visual-Propagation Decoupling Analysis}
\label{sec:exp_visual_decoupling}

\subsubsection{Dataset-wise Visual-Exit Behavior}
\label{sec:exp_exit_distribution}

Figure~\ref{fig:savb_exit_distribution} reports DecoMoE's visual-exit
results. On MMBench, 77.3\% of cases exit in layers 18--27; on
HallusionBench, 72.2\% exit in layers 28--37; and POPE exhibits a broader
distribution across multiple intervals. These dataset-dependent
distributions indicate that SAVB adapts its exit decisions instead of
relying on a fixed global boundary. In addition, 34.2\% of ScienceQA
cases do not trigger, indicating that visual tokens are retained when
the gate criterion is not satisfied.

\begin{table*}[t]
\centering
\scriptsize
\setlength{\tabcolsep}{3.0pt}

\begin{minipage}[t]{0.48\textwidth}
\centering
\textbf{(a) Expert-order comparison}\\[3pt]

\renewcommand{\arraystretch}{1.08}

\resizebox{\linewidth}{!}{%
\begin{tabular}{l|ccccc}
\toprule
\multirow[c]{2}{*}{Order Mode}
& \multicolumn{5}{c}{Keep Ratio} \\[-1.5pt]
& 10\% & 20\% & 30\% & 40\% & 50\% \\[-0.5pt]
\midrule
Native
& 87.3\% & 90.9\% & 97.0\% & 96.7\% & 97.4\% \\
Random
& 85.4\% & 89.8\% & 94.9\% & 95.5\% & 95.9\% \\
RCEP
& \textbf{96.9\%}
& \textbf{97.4\%}
& \textbf{97.7\%}
& \textbf{97.9\%}
& \textbf{97.5\%} \\
\bottomrule
\end{tabular}%
}
\end{minipage}
\hfill
\begin{minipage}[t]{0.48\textwidth}
\centering
\textbf{(b) Routed-mass target sensitivity}\\[3pt]

\renewcommand{\arraystretch}{1.45}

\resizebox{\linewidth}{!}{%
\begin{tabular}{l|cccccc}
\toprule
\multirow[c]{2}{*}{Metric}
& \multicolumn{6}{c}{Mass Target $p$} \\[-1.5pt]
& 0.1 & 0.3 & 0.5 & 0.7 & 0.9 & 1.0 \\[-0.5pt]
\midrule
Avg. Ret.
& 96.7\%
& 96.7\%
& 97.2\%
& \textbf{97.8\%}
& 97.6\%
& 97.3\% \\
Avg. Prefix
& 8.0
& 8.7
& 17.0
& 35.0
& 67.0
& 128.0 \\
\bottomrule
\end{tabular}%
}
\end{minipage}

\renewcommand{\arraystretch}{1.0}

\caption{RCEP analysis on Qwen3-VL-MoE over POPE, HallusionBench, and AI2D.
(a) Performance retention under different expert orders and fixed keep ratios.
(b) Sensitivity to the routed-mass target $p$, where Avg. Prefix denotes the
mean retained expert-prefix length.}
\label{tab:rcep_analysis}
\end{table*}

\begin{figure*}[t]
    \centering

    \begin{minipage}[t]{0.48\textwidth}
        \centering
        \includegraphics[width=\linewidth]
        {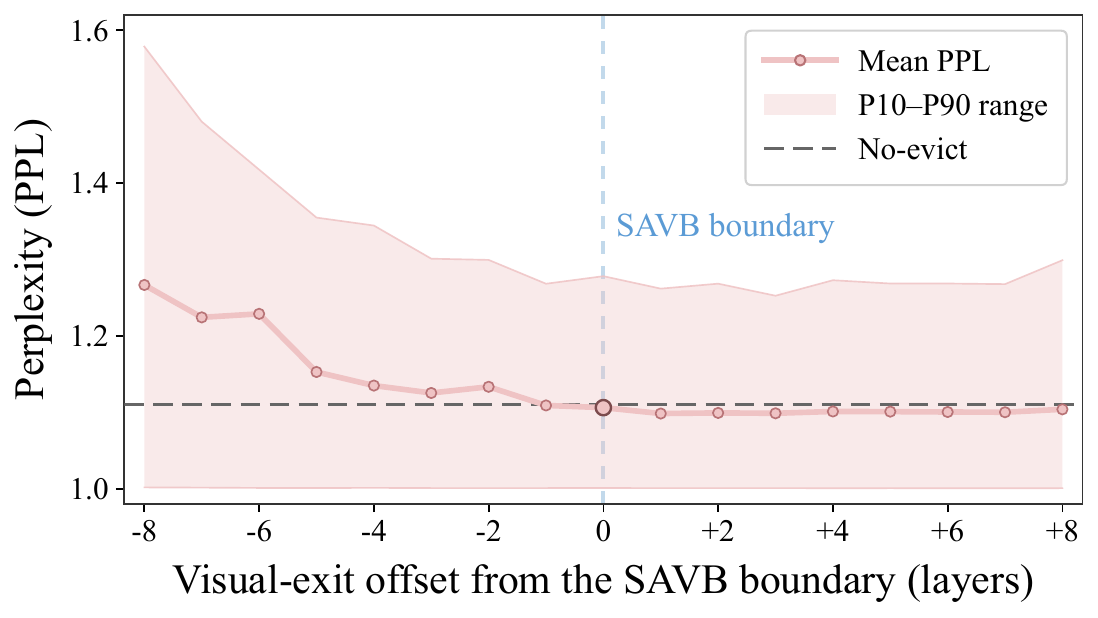}
        
        \vspace{2pt}
        \textbf{(a) Visual-propagation boundary validation}
    \end{minipage}
    \hfill
    \begin{minipage}[t]{0.48\textwidth}
        \centering
        \includegraphics[width=\linewidth]
        {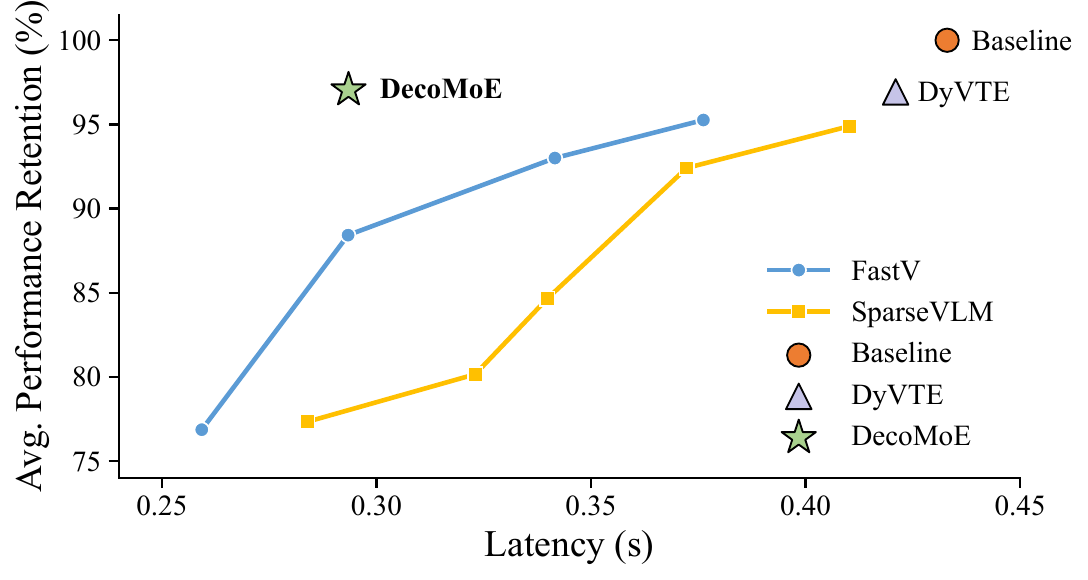}
        
        \vspace{2pt}
        \textbf{(b) Latency--performance trade-off}
    \end{minipage}

    \caption{
    Analysis of DecoMoE's visual boundary and realized acceleration.
    (a) Validation of the visual-propagation boundary predicted by SAVB,
    where visual exit is forced at offsets ranging from 8 layers before
    to 8 layers after the predicted boundary.
    (b) Latency--performance trade-off among representative efficient
    inference methods on Qwen3-VL-MoE over AI2D and HallusionBench;
    higher retention and lower latency lie toward the upper-left region.
    }
    \label{fig:savb_latency_analysis}
\end{figure*}

\subsubsection{Visual-Propagation Decoupling Validation}
\label{sec:exp_boundary_perturbation}

Figure~\ref{fig:savb_latency_analysis}(a) examines whether SAVB
identifies when visual propagation can be terminated. Exiting before the
predicted boundary increases perplexity, reaching 1.2666 at offset $-8$.
At the predicted boundary, perplexity is 1.1060, close to the
no-exit result of 1.1105, while later offsets remain within
1.0983--1.1038. These results support the view that the SAVB boundary marks a point after which visual-token removal yields PPL comparable to full visual propagation.

\subsection{Expert-Importance Decoupling Analysis}
\label{sec:exp_expert_decoupling}
\subsubsection{Mass--Performance Validation of Expert Reordering}
\label{sec:exp_rcep_mass_performance}

We evaluate whether routed mass provides a criterion for offline expert reordering by comparing the native order, a random permutation, and the RCEP order under fixed expert keep ratios. RCEP achieves the highest retention at every tested budget: at a 10\% keep ratio, it retains 96.9\% of baseline, compared with 87.3\% for Native and 85.4\% for Random. Averaged across the five budgets, RCEP reaches 97.5\%, versus 93.9\% and 92.3\%, respectively. These results indicate that Mean-Mass better distinguishes experts that are important for text-token computation and organizes them into a compact prefix, supporting the offline reordering used for expert-importance decoupling.

\subsubsection{Top-$p$ Sensitivity and Dynamic Prefix Selection}
\label{sec:exp_rcep_top_p}

Table~\ref{tab:rcep_analysis}(a) shows that retention changes little once the keep ratio reaches 30\%, indicating that a fixed large expert budget may be unnecessary. Meanwhile, Figures~\ref{fig:decoupling-evidence}(b) and~\ref{fig:decoupling-evidence}(c) show that routed-mass concentration and expert overlap vary across layers, making a fixed prefix size potentially suboptimal across layer invocations. RCEP therefore retains the shortest prefix covering target mass $p$ rather than a fixed number of experts. Table~\ref{tab:rcep_analysis}(b) shows that $p=0.70$ achieves the highest retention of 97.8\% with an average prefix of 35.0 experts.

\begin{table*}[t]
\centering
\small
\setlength{\tabcolsep}{5pt}
\renewcommand{\arraystretch}{1.10}
\begin{tabular}{l|cccccc}
\toprule
Method
& Baseline
& FastV
& SparseVLM
& DyVTE
& FastMMoE
& DecoMoE \\
\midrule
TTFT (ms)
& 370.80
& 320.77
& 339.13
& 358.93
& 1515.57
& \textbf{256.53} \\
TPOT (ms/token)
& 135.80
& 117.48
& 124.20
& 131.45
& 555.05
& \textbf{93.95} \\
\bottomrule
\end{tabular}
\caption{TTFT and TPOT comparison on Qwen3-VL-MoE. Lower is better.}
\label{tab:ttft_tpot}
\end{table*}

\subsection{Structured Execution for Realized Acceleration}
\label{sec:exp_structured_acceleration}

Figure~\ref{fig:savb_latency_analysis}(b) reports a focused latency sweep on AI2D and HallusionBench, where DecoMoE achieves 97.05\% retention at 0.29 seconds. Compared with the similar-FLOPs results in Table~\ref{tab:qwen_selected_method_efficiency}, competing methods exhibit lower performance retention at latency settings closer to that of DecoMoE. Table~\ref{tab:ttft_tpot} further shows that DecoMoE reduces TTFT from 370.80 to 256.53 ms and TPOT from 135.80 to 93.95 ms/token relative to the dense baseline, corresponding to reductions of 30.8\% in both metrics, while also achieving lower TTFT and TPOT than the compared efficient inference methods. These gains are consistent with DecoMoE's structured execution: SAVB removes a contiguous visual-token block, while RCEP restricts expert dispatch and MLP computation to a contiguous expert prefix. Such structured reductions reduce irregular selection and fragmented expert dispatch, helping theoretical computation savings translate into lower realized inference latency.

%% file: sec/5_Conclusion.tex
\section{Conclusion}
\label{sec:conclusion}

We present DecoMoE, a two-dimensional structured compression framework for multimodal MoE inference. Its central principle is to separate two decisions that are coupled in standard execution but exhibit different regularities. Visual-propagation decoupling identifies an input-dependent boundary after which visual-token removal yields PPL comparable to full visual propagation. Expert-importance decoupling separates expert importance for subsequent text-token computation from the original physical expert order, representing recurrent routing structure as a compact contiguous prefix. Together, the two mechanisms provide structured reductions along the token and expert dimensions. On Qwen3-VL-MoE, DecoMoE achieves a 1.69$\times$ measured speedup with a 2.09\% average performance reduction relative to the dense baseline.

%% file: sec/Appendix.tex
\raggedbottom

\section{Appendix}
\label{app:appendix}

This appendix provides implementation-level details and additional evidence for
the two decoupled components introduced in the main paper: the
Sample-Adaptive Visual Boundary (SAVB) and the Routing-Calibrated Expert Prefix
(RCEP). We first profile the inference bottleneck, then specify the exact offline
and online algorithms, reproduce the analytical FLOPs accounting used in our
experiments, document the calibration protocol, and finally present additional
routing diagnostics and case studies.

\subsection{Inference Bottleneck Analysis}
\label{app:bottleneck}

Figure~\ref{fig:method_and_latency}(a) reports a layer-wise end-to-end latency
breakdown. Attention is the largest component in most layers, typically
contributing approximately 57--68\% of the measured time, while the expert MLP
contributes another approximately 18--26\%. Their combined contribution
therefore generally exceeds 80\% after excluding startup and boundary effects.
This profile motivates optimizing both axes rather than treating visual-token
pruning and MoE computation as independent alternatives.

High-resolution images introduce a long visual prefix. For a prefill sequence of
length $S$, self-attention contains an $O(S^2)$ interaction term, so retaining
all visual tokens through all decoder layers repeatedly pays the quadratic
attention cost. The same visual tokens also enter each MoE block. As shown later
in Appendix~\ref{app:routing-evidence}, their routing mass is spread across a
large fraction of the expert pool, which forces the dense batched expert
implementation to maintain a large effective expert set. SAVB shortens the
sequence after the visual information has been transferred to the text states;
RCEP then reduces the physically evaluated expert prefix for the remaining
text-only computation.

\subsection{Exact DecoMoE Algorithm}
\label{app:algorithm}

\subsubsection{Offline Calibration}

Algorithm~\ref{alg:app-offline} expands the training description in the main
paper. The backbone is frozen throughout. SAVB labels are obtained by forced
visual exit at the sparse candidate layers
$\mathcal{G}=\{3,7,11,\ldots,47\}$ (zero-based implementation indices).
RCEP is calibrated separately from text-token router statistics after visual
exit. The resulting permutation is applied consistently to the router output
rows and the corresponding expert tensors; therefore reordering changes only
physical storage, not the function of the unsliced model.

\begin{algorithm*}[t]
\caption{Offline SAVB and RCEP calibration}
\label{alg:app-offline}
\small
\begin{algorithmic}[1]
\REQUIRE Frozen multimodal MoE $f$, SAVB calibration cases
$\mathcal{C}_{S}$, RCEP calibration cases $\mathcal{C}_{R}$, candidate layers
$\mathcal{G}$, NLL percentile $P_{\mathrm{NLL}}$
\ENSURE Gate parameters $\phi$ and layer-wise expert permutations
$\{\pi_l\}_{l=1}^{L}$
\FOR{each case $c\in\mathcal{C}_{S}$}
    \STATE Generate the no-exit reference answer.
    \FOR{each $l\in\mathcal{G}$}
        \STATE Force visual exit after layer $l$ without expert slicing.
        \STATE Record correctness $\kappa_{c,l}$, first-token NLL
        $\eta_{c,l}$, and feature
        $x_{c,l}=[\bar T^{(l)};T_n^{(l)}]$.
    \ENDFOR
    \STATE Set $l_c^\star=\min\{l\in\mathcal{G}:\kappa_{c,l}=1\}$ when it
    exists.
\ENDFOR
\STATE Compute $\tau(P_{\mathrm{NLL}})$ from
$\{\eta_{c,l_c^\star}\}$ and set
$z_{c,l}=\mathbb{I}[\kappa_{c,l}=1\wedge
\eta_{c,l}\leq\tau(P_{\mathrm{NLL}})]$.
\STATE Train the two-class MLP $g_\phi$ on $(x_{c,l},z_{c,l})$.
\FOR{each case $c\in\mathcal{C}_{R}$ and decoder layer $l$}
    \STATE Collect text-token routed mass $M_{c,l,e}$ after the calibrated
    visual-exit condition.
\ENDFOR
\FOR{each decoder layer $l$}
    \STATE Sort experts by descending case-mean routed mass to obtain $\pi_l$.
    \STATE Apply $\pi_l$ to router rows, gate/up expert tensors, and down
    expert tensors using the same permutation.
\ENDFOR
\end{algorithmic}
\end{algorithm*}

\subsubsection{Online Inference and Prefix-Restricted Routing}

Algorithm~\ref{alg:app-online} removes an ambiguity that is easy to miss in a
high-level description of RCEP. At a text-only MoE layer, DecoMoE first computes
router logits over all $E_l$ reordered experts. These full logits are used only
to estimate the required prefix length $K_l$. DecoMoE then slices the logits to
the first $K_l$ entries, applies a \emph{new} softmax over this prefix, reruns
native Top-$k_{\mathrm{moe}}$ inside the prefix, and renormalizes the selected
weights. It does not intersect the full-pool Top-k set with the prefix. Because
$K_l\geq k_{\mathrm{moe}}$, prefix routing always returns exactly
$k_{\mathrm{moe}}$ experts per token.

\begin{algorithm*}[t]
\caption{DecoMoE online prefill and recurrent decoding}
\label{alg:app-online}
\small
\begin{algorithmic}[1]
\REQUIRE Reordered model, gate $g_\phi$, target routed-mass fraction $p$,
native routing width $k_{\mathrm{moe}}$
\STATE Set $\mathrm{exited}\leftarrow\mathrm{false}$.
\FOR{$l=1,\ldots,L$}
    \IF{$\mathrm{exited}=\mathrm{false}$}
        \STATE Evaluate layer $l$ on $H_l=[V^{(l)};T^{(l)}]$ with the full
        expert pool.
        \IF{$l\in\mathcal{G}$ and
        $\operatorname{softmax}(g_\phi(x_l))_1\geq\theta$}
            \STATE Remove the complete visual block after layer $l$.
            \STATE Set $\mathrm{exited}\leftarrow\mathrm{true}$.
        \ENDIF
    \ELSE
        \STATE Compute full reordered router logits
        $a_{i,1:E_l}$ and probabilities $q_{i,1:E_l}$.
        \STATE Compute full-pool native Top-$k_{\mathrm{moe}}$ and aggregate
        activated mass $s_{l,e}(x)$.
        \STATE Find the smallest prefix covering $p$ of the total activated
        mass and set
        $K_l=\max(k_{\mathrm{moe}},\widehat K_l)$.
        \STATE Recompute
        $\widetilde q_{i,1:K_l}=\operatorname{softmax}(a_{i,1:K_l})$.
        \STATE Rerun Top-$k_{\mathrm{moe}}$ on $\widetilde q_{i,1:K_l}$ and
        renormalize the selected weights.
        \STATE Evaluate and combine only the first $K_l$ physically contiguous
        expert MLPs.
    \ENDIF
\ENDFOR
\STATE Reuse the shortened text-only cache and repeat the same
prefix-restricted routing during autoregressive decoding.
\end{algorithmic}
\end{algorithm*}

\paragraph{Layer-index convention.}
The implementation records zero-based layer indices. A trigger recorded as
$l$ means that layer $l$ itself is evaluated with the full multimodal sequence;
the visual block is removed from the hidden state and KV cache before layer
$l+1$. Consequently, a reported trigger at layer 27 corresponds to preserving
visual tokens through the 28th decoder block.

\subsection{Analytical FLOPs Accounting}
\label{app:flops}

We report decoder-side theoretical FLOPs using the same case-level replay code
as the main experiments. Let $L$ be the number of decoder layers, $E$ the full
expert count, $k_{\mathrm{moe}}$ the native per-token routing width, $d$ the
hidden size, $q$ the attention output size, $k_v$ the concatenated key/value
projection size used by the implementation, and $m$ the intermediate width of
one expert. A multiply-add is counted as two FLOPs. The per-token linear
projection, router, and retained-expert costs are
\begin{align}
    C_{\mathrm{proj}}
    &=2(dq+2dk_v+qd),\\
    C_{\mathrm{router}}
    &=2dE,\\
    C_{\mathrm{expert}}(K)
    &=6Kdm.
    \label{eq:app-component-flops}
\end{align}
The factor six in $C_{\mathrm{expert}}$ accounts for the gate, up, and down
matrix multiplications. For a prefill sequence of length $S$, the cost of one
decoder layer is
\begin{equation}
    \begin{aligned}
    F_{\mathrm{pre}}^{(l)}(S,K)
    &=
    S C_{\mathrm{proj}}+2qS(S+1)\\
    &\quad
    +S\!\left(C_{\mathrm{router}}+C_{\mathrm{expert}}(K)\right).
    \end{aligned}
    \label{eq:app-prefill-flops}
\end{equation}
For one decode forward with KV-cache length $C$, the corresponding layer cost is
\begin{equation}
    F_{\mathrm{dec}}^{(l)}(C,K)
    =
    C_{\mathrm{proj}}
    +C_{\mathrm{router}}
    +C_{\mathrm{expert}}(K)
    +4qC.
    \label{eq:app-decode-flops}
\end{equation}

For Qwen3-VL-MoE-30B-A3B, the replay uses
$L=48$, $E=128$, $k_{\mathrm{moe}}=8$, $d=2048$, $q=4096$,
$k_v=512$, and $m=768$. Let $S_c$, $V_c$, and $T_c=S_c-V_c$
denote the measured prefill, visual, and text-token counts for case $c$.
Let $G_c$ be its generated-token count. Prefill produces the first generated
token, so the number of recurrent decode forwards is
$D_c=\max(G_c-1,0)$.

If SAVB triggers after layer $l_c^\star$, the case-level DecoMoE cost is replayed
as
\begin{align}
F^{c}_{\mathrm{pre}}
&=
\sum_{l=1}^{l_c^\star+1}
F_{\mathrm{pre}}^{(l)}(S_c,E)
+
\sum_{l=l_c^\star+2}^{L}
F_{\mathrm{pre}}^{(l)}(T_c,K^c_l),\\
F^{c}_{\mathrm{dec}}
&=
\sum_{t=1}^{D_c}\sum_{l=1}^{L}
F_{\mathrm{dec}}^{(l)}(T_c+t-1,K^c_{l,t}).
\label{eq:app-decomoe-case-flops}
\end{align}
Here $K^c_l$ and $K^c_{l,t}$ are read from the recorded prefill and decode RCEP
traces. If a trace entry is unavailable, the replay uses the case-level mean
retained expert count. If SAVB never triggers, all layers use $(S_c,E)$.

The SAVB MLP uses an input of width $2d$, hidden width $h=512$, and a two-class
output. If it is evaluated $n_c^{\mathrm{gate}}$ times, its explicitly counted
overhead is
\begin{equation}
    F^c_{\mathrm{gate}}
    =
    n_c^{\mathrm{gate}}
    \left(2(2d)h+2h\cdot2\right).
    \label{eq:app-gate-flops}
\end{equation}
The final case total is
$F_c=F^c_{\mathrm{pre}}+F^c_{\mathrm{dec}}+F^c_{\mathrm{gate}}$.
Reported total saving is computed from sums over the same matched cases:
\begin{equation}
    \operatorname{Saving}
    =
    1-\frac{\sum_c F_c^{\mathrm{method}}}
            {\sum_c F_c^{\mathrm{baseline}}}.
    \label{eq:app-saving}
\end{equation}

\paragraph{Dense-equivalent expert convention.}
The evaluated Qwen inference path forms a dense batched BMM over every
physically retained expert and applies sparse routing weights afterward.
Accordingly, the baseline expert term uses $K=E=128$, whereas RCEP uses the
recorded prefix length $K_l$. This convention follows executed expert matrix
multiplications rather than counting only the eight nonzero routing weights.

\paragraph{Scope and limitations.}
These equations count the language decoder backbone and the explicit SAVB MLP.
They do not include the vision encoder, embeddings, normalization, LM head,
sampling, Python control flow, softmax/Top-k/cumulative-sum kernels, memory
movement, or kernel-launch overhead. Full router projection over all $E$ experts
is included even after RCEP activates. Therefore theoretical FLOPs and measured
profiler CUDA time are complementary quantities; neither is inferred from the
other.

\subsection{Calibration Data and Data Isolation}
\label{app:calibration}

\subsubsection{SAVB Supervision}

The default Qwen SAVB checkpoint is trained from direct-answer forced-exit
labels collected on POPE and MME. We randomly sample 50\% of each benchmark
using a fixed seed and evaluate visual-only eviction with all 128 experts kept,
so expert compression cannot contaminate the boundary labels. Candidate layers
are $3,7,11,\ldots,47$. The default $P95$ setting uses the 95th percentile of
the global first-good-layer NLL distribution to construct the independent
case--layer labels defined in the main paper. The default model uses an
independent two-class MLP at each candidate position, with a 512-dimensional
hidden layer and cross-entropy training on frozen features. The reported
checkpoint is taken after epoch 2.

\subsubsection{RCEP Calibration and Text-Routing Alignment}

RCEP uses an independent calibration pass over a random 10\% sample from POPE
and MME with direct-answer prompts. The current scripts do not enforce this
sample to be disjoint from the SAVB sample. During collection, visual tokens are
removed after implementation layer 24 and no expert slicing is enabled. Routed
mass is computed only from text-token positions. We average each expert's
routed mass first within a case and then across cases, which prevents long
examples from dominating the layer-wise order. The same permutation is applied
to the router output rows and both expert parameter tensors before evaluation.

The fixed layer-24 collection condition is used to expose text-aligned routing,
not to constrain online execution. RCEP reorders every calibrated MoE layer,
including layers before 24, while online prefix slicing starts only after the
sample-specific SAVB trigger. Figure~\ref{fig:app-routing-consistency} and
Figure~\ref{fig:app-mass-stability} verify that removing visual tokens does
not materially change the remaining text-token routing distribution in deep
layers, supporting transfer of the calibrated order to dynamic SAVB exits.

\subsubsection{Leakage Boundary and Reporting Protocol}

The calibration procedures never update the multimodal backbone and never use
examples, labels, or answers from MMBench, ScienceQA, AI2D, AOKVQA, MMMU, or
HallusionBench. Results on these six datasets are therefore strict
cross-dataset generalization tests. The current experiment scripts draw SAVB
and RCEP calibration subsets from the POPE/MME benchmark pools themselves.
Consequently, aggregate POPE/MME numbers that include those calibration IDs
should be described as \emph{transductive calibration results}, not as a strict
held-out test.

For a fully inductive POPE/MME report, calibration IDs must be persisted before
training and excluded from evaluation; only the complementary IDs may be
scored. This distinction does not affect the six unseen datasets, but it is
important for reproducibility and prevents calibration overlap from being
mischaracterized as test-set generalization.

\subsection{Complete RCEP Sensitivity Results}
\label{app:rcep-complete-results}

The main paper reports a compact subset of the RCEP sensitivity results to
reduce space. Here we provide the complete expert-order comparison and
routed-mass target sweep used in the original analysis.

\subsubsection{Complete Expert-Order Comparison}

Table~\ref{tab:app-rcep-mass-performance} reports all seven evaluated expert
keep ratios. RCEP achieves the highest retention at every tested budget. In
particular, at a 10\% keep ratio it retains 96.87\% of baseline performance,
compared with 87.26\% for the native order and 85.43\% for a random order.
Across all seven budgets, the corresponding mean retentions are 97.47\%,
93.02\%, and 92.56\% for RCEP, Native, and Random, respectively. These complete
results support the same conclusion as the compact table in the main paper:
text-routed mass provides a stable offline signal for organizing important
experts into a contiguous prefix.

\begin{table}[t]
\centering
\scriptsize
\setlength{\tabcolsep}{2.6pt}
\renewcommand{\arraystretch}{1.08}
\resizebox{\columnwidth}{!}{%
\begin{tabular}{l|ccccccc}
\toprule
\multirow[c]{2}{*}{Order Mode}
& \multicolumn{7}{c}{Keep Ratio} \\[-1.5pt]
& 10\% & 15\% & 20\% & 25\% & 30\% & 40\% & 50\% \\[-0.5pt]
\midrule
Native
& 87.26\% & 86.60\% & 90.85\% & 95.39\%
& 96.97\% & 96.67\% & 97.41\% \\
Random
& 85.43\% & 94.77\% & 89.79\% & 91.57\%
& 94.93\% & 95.50\% & 95.94\% \\
RCEP
& \textbf{96.87\%} & \textbf{97.39\%}
& \textbf{97.41\%} & \textbf{97.55\%}
& \textbf{97.71\%} & \textbf{97.91\%}
& \textbf{97.47\%} \\
\bottomrule
\end{tabular}%
}
\caption{Complete mass--performance comparison of Native, Random, and RCEP
expert orders on Qwen3-VL-MoE over POPE, HallusionBench, and AI2D. Entries
report Avg. Ret. under fixed expert keep ratios.}
\label{tab:app-rcep-mass-performance}
\end{table}

\subsubsection{Complete Routed-Mass Target Sensitivity}

Table~\ref{tab:app-rcep-top-p-sensitivity} restores the full sweep from
$p=0.1$ to $1.0$. Retention remains relatively stable across a broad range of
mass targets, while the retained prefix grows monotonically as $p$ increases.
The default $p=0.70$ gives the highest measured retention, 97.8\%, with an
average prefix of 35.0 experts, providing a favorable operating point between
performance preservation and expert-prefix length.

\begin{table}[t]
\centering
\scriptsize
\setlength{\tabcolsep}{2.2pt}
\renewcommand{\arraystretch}{1.0}
\resizebox{\columnwidth}{!}{%
\begin{tabular}{l|*{10}{c}}
\toprule
\multirow[c]{2}{*}{Metric}
& \multicolumn{10}{c}{Mass Target $p$} \\[-1.5pt]
& 0.1 & 0.2 & 0.3 & 0.4 & 0.5
& 0.6 & 0.7 & 0.8 & 0.9 & 1.0 \\[-0.5pt]
\midrule
\rule{0pt}{2.5ex}Avg. Ret.
& 96.7\% & 96.8\% & 96.7\% & 96.9\% & 97.2\%
& 97.6\% & \textbf{97.8\%} & 97.2\% & 97.6\% & 97.3\% \\[2pt]
\rule{0pt}{2.5ex}Avg. Prefix
& 8.0 & 8.0 & 8.7 & 11.7 & 17.0
& 24.7 & 35.0 & 49.0 & 67.0 & 128.0 \\
\bottomrule
\end{tabular}%
}
\caption{Complete sensitivity sweep for the RCEP routed-mass target $p$ on
Qwen3-VL-MoE over POPE, HallusionBench, and AI2D. Avg. Prefix is the mean
retained expert-prefix length.}
\label{tab:app-rcep-top-p-sensitivity}
\end{table}

\subsection{Additional Routed-Mass Evidence}
\label{app:routing-evidence}

\begin{figure}[t]
    \centering
    \includegraphics[width=0.98\columnwidth]{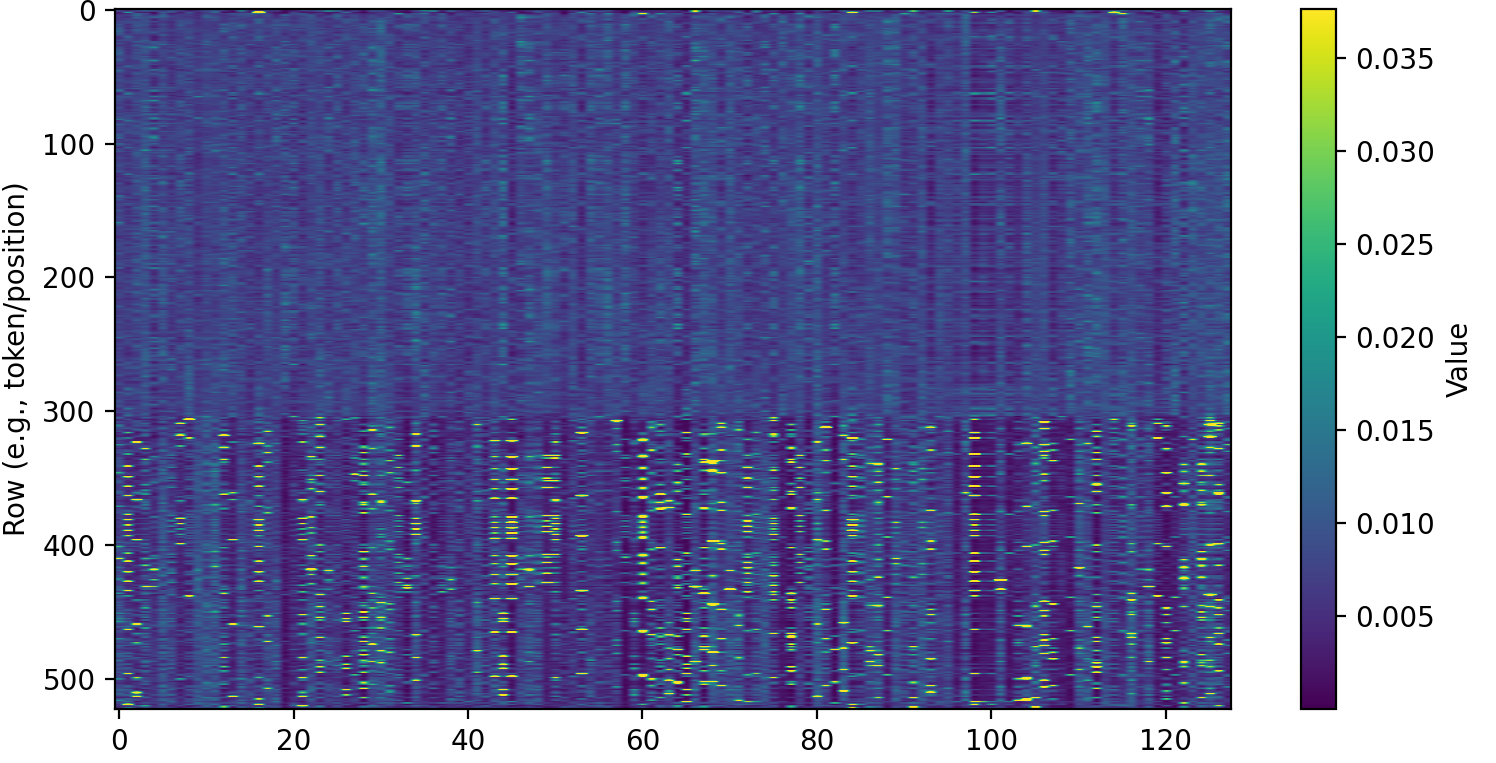}
    \caption{Token-to-expert routing map. Visual-token routing is diffuse and
    spans a broad expert set, whereas text-token routing forms a substantially
    more concentrated and recurrent structure.}
    \label{fig:app-routing-visual-text}
\end{figure}

Figure~\ref{fig:app-routing-visual-text} visualizes token-level expert
activation. Visual tokens distribute routed mass across many experts, forcing a
large effective expert set while they remain in the sequence. Text tokens form
more concentrated vertical structures, indicating that a shared text-aligned
expert order can capture recurring importance across cases.

\begin{figure}[t]
    \centering
    \includegraphics[width=\columnwidth]{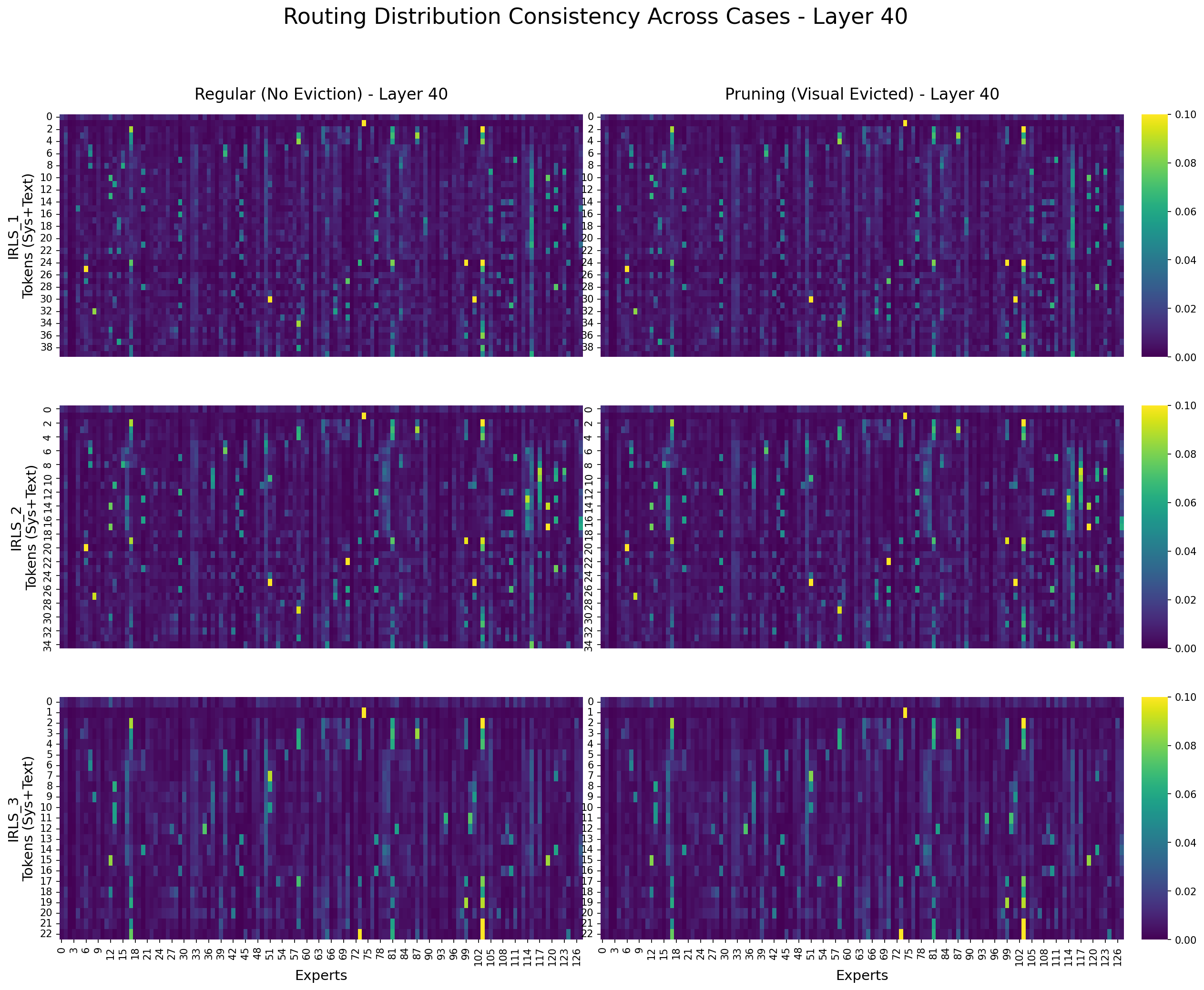}
    \caption{Text-token router distributions at layer 40 for three cases,
    before and after visual-token eviction. The two conditions retain nearly
    identical structures.}
    \label{fig:app-routing-consistency}
\end{figure}

Figure~\ref{fig:app-routing-consistency} compares the remaining text-token
router maps with and without visual-token eviction. Across cases, the dominant
experts and relative mass patterns remain visually aligned. This result is
important for the SAVB--RCEP interface: expert importance calibrated in a
text-only state remains representative after a sample-specific visual exit.

\begin{figure}[t]
    \centering
    \includegraphics[width=0.72\columnwidth]{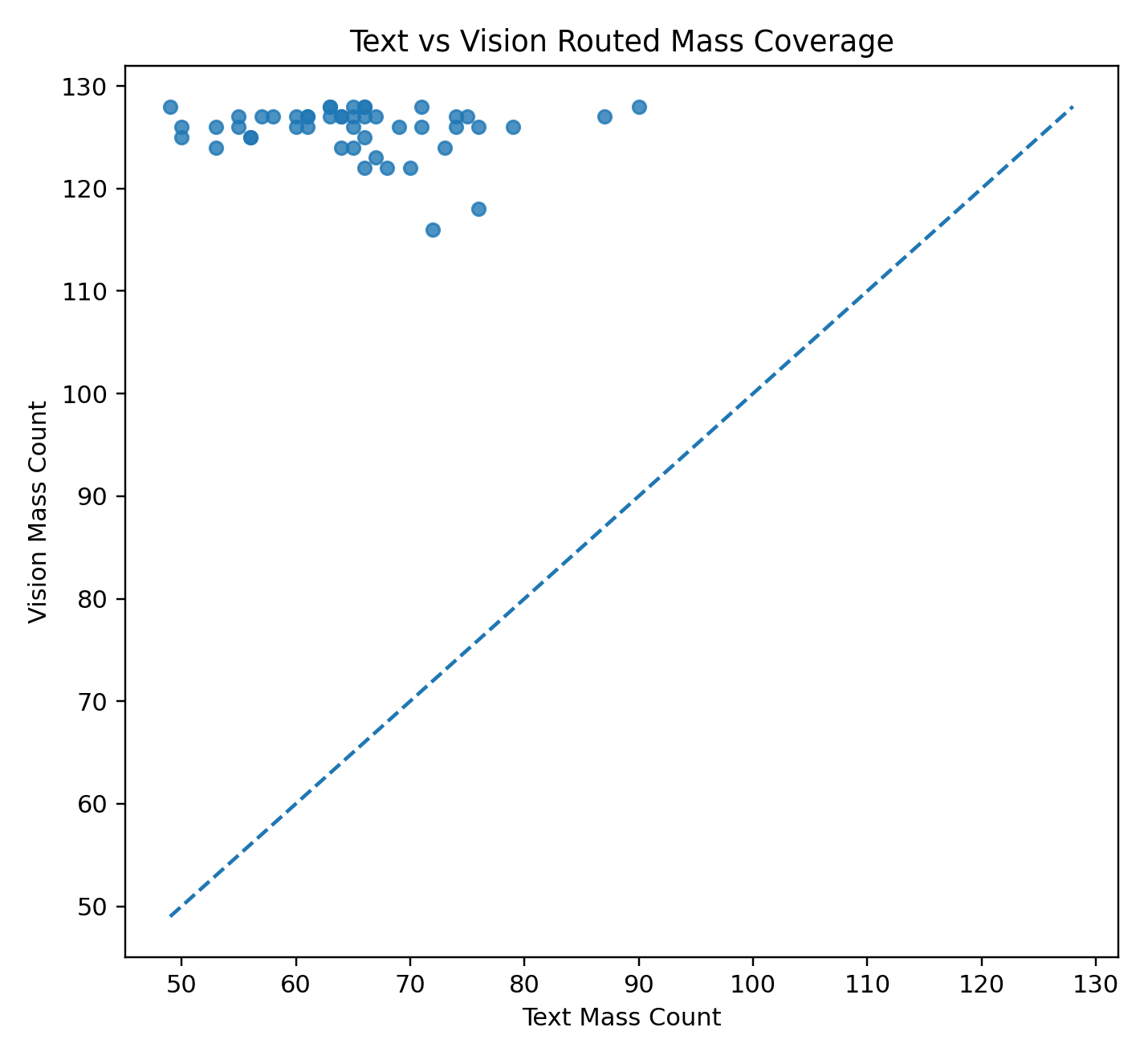}

    \vspace{0.4em}
    \includegraphics[width=\columnwidth]{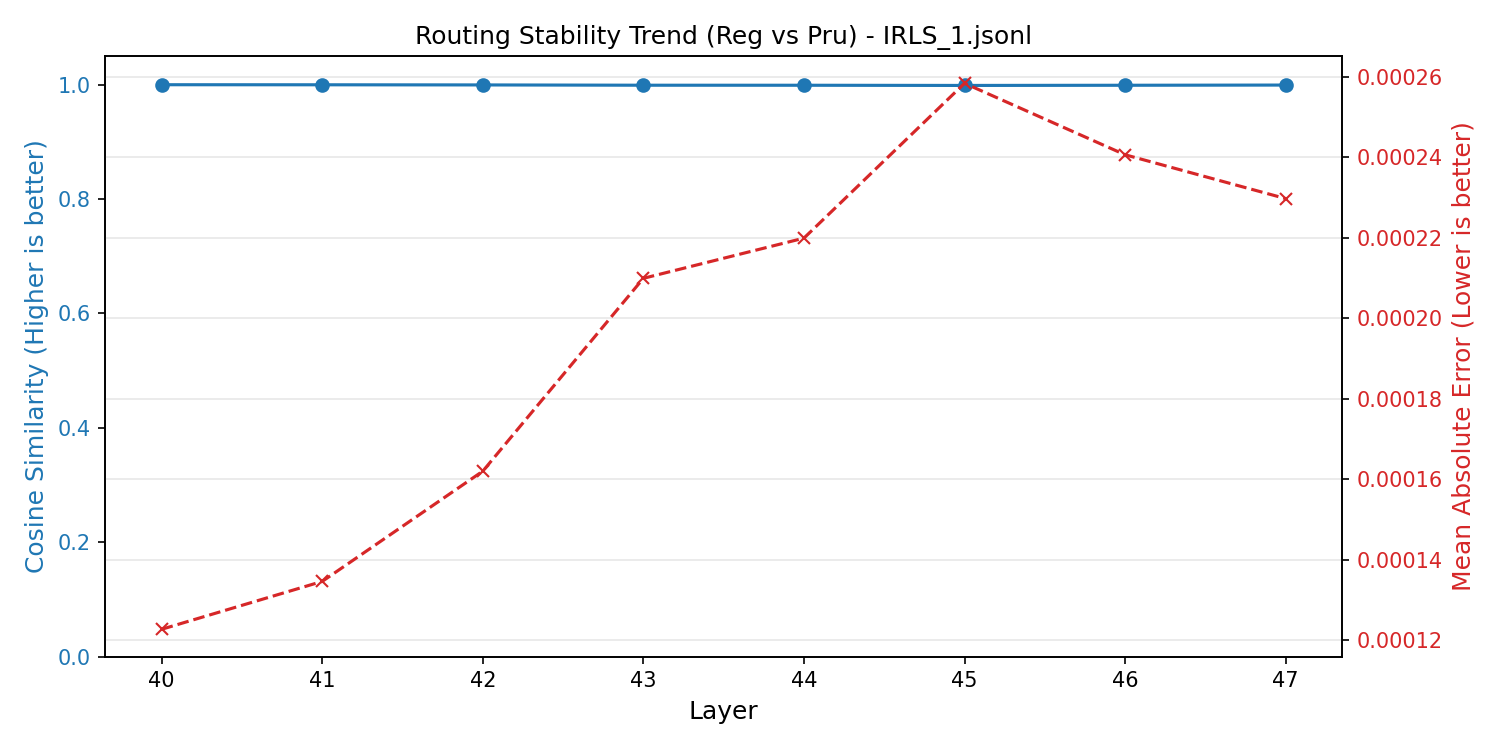}
    \caption{Complementary routed-mass diagnostics. Top: number of experts
    required to cover 50\% routed mass; most text cases require roughly
    50--80 experts, whereas visual tokens lie near 120--128 experts. Bottom:
    deep-layer text-routing stability after visual-token eviction; cosine
    similarity remains above 0.997 and MAE below 0.0003.}
    \label{fig:app-mass-stability}
\end{figure}

The mass-count scatter in the top panel of
Figure~\ref{fig:app-mass-stability} rules out a purely
superficial coverage effect. At a routed-mass threshold of 0.5, visual tokens
assign substantial probability to almost the entire 128-expert pool, with most
points near 120--128 experts. Text tokens generally require only approximately
50--80 experts, with a small number of harder cases extending beyond this
range. Thus visual propagation does not merely touch more experts with
negligible weights; it preserves broad, nontrivial expert activity.

The bottom panel of Figure~\ref{fig:app-mass-stability} quantifies the same
phenomenon over deep
layers. The router-distribution cosine similarity remains above 0.997, while
the mean absolute error remains below 0.0003. Early visual eviction therefore
does not redirect the remaining text tokens to a different family of experts;
the original high-probability text experts remain stable. Together with
the left panel, this observation explains why expert-prefix
compression should be activated \emph{after} the learned visual boundary rather
than while visual routing is still diffuse.

\subsection{Why the Visual Boundary Must Be Sample Adaptive}
\label{app:case-studies}

\subsubsection{Case 1: Many Visual Tokens but an Early Boundary}

\begin{figure}[t]
    \centering
    \includegraphics[width=0.98\columnwidth]{sec/Figs/appendix/case_many_tokens.png}
    \caption{The snowy residential scene used in Case 1. Its high-resolution
    input produces approximately 4,000 visual tokens.}
    \label{fig:app-case-many}
\end{figure}

\paragraph{Prompt.}
\begin{quote}\small
Task: Answer the following multiple-choice questions about the image, then
answer the detailed questions.

1. What type of scene is shown? (A) Indoor (B) Residential outdoor
(C) Commercial (D) Natural.\\
2. What season is depicted? (A) Spring (B) Summer (C) Autumn (D) Winter.\\
3. What recent weather event is evident? (A) Heavy rain (B) Strong wind
(C) Snowfall (D) Fog.\\
4. Which is clearly visible? (A) Vehicles (B) People (C) Buildings
(D) Animals.\\
5. The tallest trees are located where relative to the buildings?
(A) In front (B) Behind (C) On rooftops (D) Inside.\\
6. Describe the trees and how snow has accumulated on them.\\
7. Describe the background buildings, including facades, windows, balconies,
and roofs.\\
8. Identify small or subtle infrastructure or activity.\\
\textbf{9. Is there a traffic signal or parking-lot sign in the picture?}
\end{quote}

\paragraph{Answer at $K=26$.}
\begin{quote}\small
1. \textbf{(B) Residential outdoor}. 2. \textbf{(D) Winter}.
3. \textbf{(C) Snowfall}. 4. \textbf{(C) Buildings}.
5. \textbf{(A) In front}. The trees are tall evergreens and deciduous trees
whose branches carry a thick layer of fresh snow. The background contains
multi-story residential buildings with light-colored facades, repeated rows of
windows and balconies, and snow-covered roofs. Smaller details include parked
cars, fences, paths, and residential street infrastructure partly occluded by
the trees and snow. \textbf{No traffic signal or parking-lot sign is visible.}
\end{quote}

Despite the large visual-token count, removing the visual block after layer 26
preserves both fine visual details and the model's textual response ability.
The case demonstrates that a long image prefix need not be propagated to a late
layer once its information has been consolidated into text states.

\subsubsection{Case 2: Few Visual Tokens but a Late Boundary}

\begin{figure}[t]
    \centering
    \includegraphics[width=0.48\columnwidth]{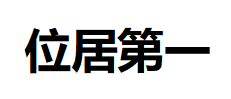}
    \caption{The compact OCR input used in Case 2, containing approximately
    80 visual tokens.}
    \label{fig:app-case-few}
\end{figure}

\paragraph{Prompt.}
\begin{quote}\small
1. What kind of image is this? 2. From which country is the text?
3. How many words or characters are shown? 4. What is the English translation?
5. What color is the text?
\end{quote}

\paragraph{Answer with exit at $K=42$.}
\begin{quote}\small
This is a digital image containing a four-character Chinese phrase. The phrase
is read as \emph{wei ju di yi} and means \textbf{``ranked first''} or
\textbf{``in first place''}. The text is black on a white background.
\end{quote}

\paragraph{Answer with exit at $K=41$.}
\begin{quote}\small
The response misreads and reorders the four characters, gives an incorrect
translation, and then enters a long repetitive explanation
\emph{(repetitive tail omitted)}.
\end{quote}

Unlike Case 1, this image is visually small but semantically precise. Exiting
one layer too early causes a discontinuous OCR failure, whereas exiting at
layer 42 preserves the correct phrase and translation. Token count is therefore
an insufficient boundary heuristic; the required depth depends on when the
task-relevant visual semantics become available to the text stream.

\subsubsection{Case 3: The Same Image Requires Different Depths by Task}

\begin{figure}[t]
    \centering
    \includegraphics[width=0.62\columnwidth]{sec/Figs/appendix/case_task_depth.png}
    \caption{The fire-hydrant image shared by the three prompts in Case 3.}
    \label{fig:app-case-task-depth}
\end{figure}

\paragraph{Level 1: Seen ($K=18$).}
\begin{quote}\small
\textbf{Prompt:} Is there a blue and yellow fire hydrant in the image? Answer
the question using a single word or phrase.\\
\textbf{Answer:} Yes.
\end{quote}

\paragraph{Level 2: Inspect ($K=22$).}
\begin{quote}\small
\textbf{Prompt:} Is there a blue and yellow fire hydrant in the image? Why?\\
\textbf{Answer:} Yes. A fire hydrant is clearly visible beside the road; its
body is blue and its top and side caps are yellow.
\end{quote}

\paragraph{Level 3: Understand ($K=42$).}
\begin{quote}\small
\textbf{Prompt:} Identify the object and its colors, specify where it is
located, describe its physical features and surrounding environment, and read
its identification number.\\
\textbf{Answer:} The object is a fire hydrant with a blue body and yellow top
and caps. It stands on the sidewalk beside a road. The hydrant has a cylindrical
body, a domed top, side outlets, and metal fittings; the surrounding scene is an
urban roadside area. Its identification number is \textbf{459775}.
\end{quote}

All three prompts use identical visual evidence, yet their required
cross-modal reasoning depths differ substantially. Object presence is available
early, grounded color justification requires additional propagation, and
fine-grained attribute extraction requires a much later boundary. This
controlled case supports SAVB's central design choice: visual exit must depend
on the joint image--prompt state rather than image resolution, visual-token
count, or a globally fixed layer.

%% file: iclr2027_conference.bib
@article{wu2026accelerating,
  title={Accelerating multimodal large language models via dynamic visual-token exit and the empirical findings},
  author={Wu, Qiong and Lin, Wenhao and Zhou, Yiyi and Ye, Weihao and Zeng, Zhanpeng and Sun, Xiaoshuai and Ji, Rongrong},
  journal={Advances in Neural Information Processing Systems},
  volume={38},
  pages={168378--168403},
  year={2026}
}

@article{liu2023visual,
  title={Visual instruction tuning},
  author={Liu, Haotian and Li, Chunyuan and Wu, Qingyang and Lee, Yong Jae},
  journal={Advances in neural information processing systems},
  volume={36},
  pages={34892--34916},
  year={2023}
}

@article{bai2023qwenvl,
  title   = {{Qwen-VL}: A Frontier Large Vision-Language Model with Versatile Abilities},
  author  = {Bai, Jinze and Bai, Shuai and Yang, Shusheng and Wang, Shijie and Tan, Sinan and Wang, Peng and Lin, Junyang and Zhou, Chang and Zhou, Jingren},
  journal = {arXiv preprint arXiv:2308.12966},
  year    = {2023},
  url     = {https://arxiv.org/abs/2308.12966}
}

@article{wang2024qwen2vl,
  title={Qwen2-vl: Enhancing vision-language model's perception of the world at any resolution},
  author={Wang, Peng and Bai, Shuai and Tan, Sinan and Wang, Shijie and Fan, Zhihao and Bai, Jinze and Chen, Keqin and Liu, Xuejing and Wang, Jialin and Ge, Wenbin and others},
  journal={arXiv preprint arXiv:2409.12191},
  year={2024}
}

@inproceedings{chen2024internvl,
  title={Internvl: Scaling up vision foundation models and aligning for generic visual-linguistic tasks},
  author={Chen, Zhe and Wu, Jiannan and Wang, Wenhai and Su, Weijie and Chen, Guo and Xing, Sen and Zhong, Muyan and Zhang, Qinglong and Zhu, Xizhou and Lu, Lewei and others},
  booktitle={Proceedings of the IEEE/CVF conference on computer vision and pattern recognition},
  pages={24185--24198},
  year={2024}
}

@article{jiang2024mixtral,
  title={Mixtral of experts},
  author={Jiang, Albert Q and Sablayrolles, Alexandre and Roux, Antoine and Mensch, Arthur and Savary, Blanche and Bamford, Chris and Chaplot, Devendra Singh and Casas, Diego de las and Hanna, Emma Bou and Bressand, Florian and others},
  journal={arXiv preprint arXiv:2401.04088},
  year={2024}
}

@inproceedings{dai2024deepseekmoe,
  title={Deepseekmoe: Towards ultimate expert specialization in mixture-of-experts language models},
  author={Dai, Damai and Deng, Chengqi and Zhao, Chenggang and Xu, RX and Gao, Huazuo and Chen, Deli and Li, Jiashi and Zeng, Wangding and Yu, Xingkai and Wu, Yu and others},
  booktitle={Proceedings of the 62nd Annual Meeting of the Association for Computational Linguistics (Volume 1: Long Papers)},
  pages={1280--1297},
  year={2024}
}

@article{lin2024moellava,
  title={Moe-llava: Mixture of experts for large vision-language models},
  author={Lin, Bin and Tang, Zhenyu and Ye, Yang and Huang, Jinfa and Zhang, Junwu and Pang, Yatian and Jin, Peng and Ning, Munan and Luo, Jiebo and Yuan, Li},
  journal={IEEE Transactions on Multimedia},
  year={2026},
  publisher={IEEE}
}

@article{wu2024deepseekvl2,
  title={Deepseek-vl2: Mixture-of-experts vision-language models for advanced multimodal understanding},
  author={Wu, Zhiyu and Chen, Xiaokang and Pan, Zizheng and Liu, Xingchao and Liu, Wen and Dai, Damai and Gao, Huazuo and Ma, Yiyang and Wu, Chengyue and Wang, Bingxuan and others},
  journal={arXiv preprint arXiv:2412.10302},
  year={2024}
}

@article{kimiteam2025kimivl,
  title={Kimi-vl technical report},
  author={Team, Kimi and Du, Angang and Yin, Bohong and Xing, Bowei and Qu, Bowen and Wang, Bowen and Chen, Cheng and Zhang, Chenlin and Du, Chenzhuang and Wei, Chu and others},
  journal={arXiv preprint arXiv:2504.07491},
  year={2025}
}

@article{bai2025qwen3vl,
  title={Qwen3-vl technical report},
  author={Bai, Shuai and Cai, Yuxuan and Chen, Ruizhe and Chen, Keqin and Chen, Xionghui and Cheng, Zesen and Deng, Lianghao and Ding, Wei and Gao, Chang and Ge, Chunjiang and others},
  journal={arXiv preprint arXiv:2511.21631},
  year={2025}
}

@article{wang2025internvl35,
  title={Internvl3. 5: Advancing open-source multimodal models in versatility, reasoning, and efficiency},
  author={Wang, Weiyun and Gao, Zhangwei and Gu, Lixin and Pu, Hengjun and Cui, Long and Wei, Xingguang and Liu, Zhaoyang and Jing, Linglin and Ye, Shenglong and Shao, Jie and others},
  journal={arXiv preprint arXiv:2508.18265},
  year={2025}
}

@inproceedings{chen2024fastv,
  title={An image is worth 1/2 tokens after layer 2: Plug-and-play inference acceleration for large vision-language models},
  author={Chen, Liang and Zhao, Haozhe and Liu, Tianyu and Bai, Shuai and Lin, Junyang and Zhou, Chang and Chang, Baobao},
  booktitle={European Conference on Computer Vision},
  pages={19--35},
  year={2024},
  organization={Springer}
}

@article{zhang2025sparsevlm,
  title={Sparsevlm: Visual token sparsification for efficient vision-language model inference},
  author={Zhang, Yuan and Fan, Chun-Kai and Ma, Junpeng and Zheng, Wenzhao and Huang, Tao and Cheng, Kuan and Gudovskiy, Denis and Okuno, Tomoyuki and Nakata, Yohei and Keutzer, Kurt and others},
  journal={arXiv preprint arXiv:2410.04417},
  year={2024}
}

@inproceedings{lin2025vtw,
  title={Boosting multimodal large language models with visual tokens withdrawal for rapid inference},
  author={Lin, Zhihang and Lin, Mingbao and Lin, Luxi and Ji, Rongrong},
  booktitle={Proceedings of the AAAI Conference on Artificial Intelligence},
  volume={39},
  number={5},
  pages={5334--5342},
  year={2025}
}

@article{li2024tokenpacker,
  title={Tokenpacker: Efficient visual projector for multimodal llm},
  author={Li, Wentong and Yuan, Yuqian and Liu, Jian and Tang, Dongqi and Wang, Song and Qin, Jie and Zhu, Jianke and Zhang, Lei},
  journal={International Journal of Computer Vision},
  volume={133},
  number={10},
  pages={6794--6812},
  year={2025},
  publisher={Springer}
}

@article{shang2025llavaprumerge,
  title={Llava-prumerge: Adaptive token reduction for efficient large multimodal models},
  author={Shang, Yuzhang and Cai, Mu and Xu, Bingxin and Lee, Yong Jae and Yan, Yan},
  journal={arXiv preprint arXiv:2403.15388},
  year={2024}
}

@inproceedings{yang2025visionzip,
  title={Visionzip: Longer is better but not necessary in vision language models},
  author={Yang, Senqiao and Chen, Yukang and Tian, Zhuotao and Wang, Chengyao and Li, Jingyao and Yu, Bei and Jia, Jiaya},
  booktitle={Proceedings of the IEEE/CVF Conference on Computer Vision and Pattern Recognition},
  pages={19792--19802},
  year={2025}
}

@inproceedings{ye2024fitprune,
  title={Fit and prune: Fast and training-free visual token pruning for multi-modal large language models},
  author={Ye, Weihao and Wu, Qiong and Lin, Wenhao and Zhou, Yiyi},
  booktitle={Proceedings of the AAAI Conference on Artificial Intelligence},
  volume={39},
  number={21},
  pages={22128--22136},
  year={2025}
}

@inproceedings{xing2025pyramiddrop,
  title={Conical visual concentration for efficient large vision-language models},
  author={Xing, Long and Huang, Qidong and Dong, Xiaoyi and Lu, Jiajie and Zhang, Pan and Zang, Yuhang and Cao, Yuhang and He, Conghui and Wang, Jiaqi and Wu, Feng and others},
  booktitle={Proceedings of the IEEE/CVF Conference on Computer Vision and Pattern Recognition},
  pages={14593--14603},
  year={2025}
}

@inproceedings{zhang2025vispruner,
  title={Beyond text-visual attention: Exploiting visual cues for effective token pruning in vlms},
  author={Zhang, Qizhe and Cheng, Aosong and Lu, Ming and Zhang, Renrui and Zhuo, Zhiyong and Cao, Jiajun and Guo, Shaobo and She, Qi and Zhang, Shanghang},
  booktitle={Proceedings of the IEEE/CVF International Conference on Computer Vision},
  pages={20857--20867},
  year={2025}
}

@inproceedings{yang2025topv,
  title={Topv: Compatible token pruning with inference time optimization for fast and low-memory multimodal vision language model},
  author={Yang, Cheng and Sui, Yang and Xiao, Jinqi and Huang, Lingyi and Gong, Yu and Li, Chendi and Yan, Jinghua and Bai, Yu and Sadayappan, Ponnuswamy and Hu, Xia and others},
  booktitle={Proceedings of the Computer Vision and Pattern Recognition Conference},
  pages={19803--19813},
  year={2025}
}

@article{tan2025tokencarve,
  title={Tokencarve: Information-preserving visual token compression in multimodal large language models},
  author={Tan, Xudong and Ye, Peng and Tu, Chongjun and Cao, Jianjian and Yang, Yaoxin and Zhang, Lin and Zhou, Dongzhan and Chen, Tao},
  journal={arXiv preprint arXiv:2503.10501},
  year={2025}
}

@inproceedings{huang2025dynamicllava,
  title={Dynamic-llava: Efficient multimodal large language models via dynamic vision-language context sparsification},
  author={Huang, Wenxuan and Zhai, Zijie and Shen, Yunhang and Cao, Shaosheng and Zhao, Fei and Xu, Xiangfeng and Ye, Zheyu and Lin, Shaohui},
  booktitle={International Conference on Learning Representations},
  volume={2025},
  pages={69927--69955},
  year={2025}
}

@inproceedings{zhang2025llavamini,
  title={Llava-mini: Efficient image and video large multimodal models with one vision token},
  author={Zhang, Shaolei and Fang, Qingkai and Yang, Yang and Feng, Yang},
  booktitle={International Conference on Learning Representations},
  volume={2025},
  pages={53285--53310},
  year={2025}
}

@article{liu2024mustdrop,
  title={Multi-stage vision token dropping: Towards efficient multimodal large language model},
  author={Liu, Ting and Shi, Liangtao and Hong, Richang and Hu, Yue and Yin, Quanjun and Zhang, Linfeng},
  journal={arXiv preprint arXiv:2411.10803},
  year={2024}
}

@inproceedings{lu2024experts,
  title={Not all experts are equal: Efficient expert pruning and skipping for mixture-of-experts large language models},
  author={Lu, Xudong and Liu, Qi and Xu, Yuhui and Zhou, Aojun and Huang, Siyuan and Zhang, Bo and Yan, Junchi and Li, Hongsheng},
  booktitle={Proceedings of the 62nd Annual Meeting of the Association for Computational Linguistics (Volume 1: Long Papers)},
  pages={6159--6172},
  year={2024}
}

@article{huang2025shapleymoe,
  title={Discovering important experts for mixture-of-experts models pruning through a theoretical perspective},
  author={Huang, Weizhong and Zhang, Yuxin and Zheng, Xiawu and Chao, Fei and Ji, Rongrong and Cao, Liujuan},
  journal={Advances in Neural Information Processing Systems},
  volume={38},
  pages={135973--136003},
  year={2026}
}

@article{xia2025fastmmoe,
  title={FastMMoE: Accelerating Multimodal Large Language Models through Dynamic Expert Activation and Routing-Aware Token Pruning},
  author={Xia, Guoyang and Ding, Yifeng and Li, Fengfa and Ren, Lei and Chen, Wei and Feng, Fangxiang and Wang, Xiaojie},
  journal={arXiv preprint arXiv:2511.17885},
  year={2025}
}

@article{huang2026modes,
  title={MoDES: Accelerating Mixture-of-Experts Multimodal Large Language Models via Dynamic Expert Skipping},
  author={Huang, Yushi and Wang, Zining and Yuan, Zhihang and Ding, Yifu and Gong, Ruihao and Guo, Jinyang and Liu, Xianglong and Zhang, Jun},
  journal={arXiv preprint arXiv:2511.15690},
  year={2025}
}

@inproceedings{li2026macs,
  title={MACS: Modality-Aware Capacity Scaling for Efficient Multimodal MoE Inference},
  author={Li, Bo and Wu, Chuan and Zhu, Shaolin},
  booktitle={Proceedings of the 64th Annual Meeting of the Association for Computational Linguistics (Volume 1: Long Papers)},
  pages={22134--22148},
  year={2026}
}

@inproceedings{yang2024moei2,
  title={Moe-i2: Compressing mixture of experts models through inter-expert pruning and intra-expert low-rank decomposition},
  author={Yang, Cheng and Sui, Yang and Xiao, Jinqi and Huang, Lingyi and Gong, Yu and Duan, Yuanlin and Jia, Wenqi and Yin, Miao and Cheng, Yu and Yuan, Bo},
  booktitle={Findings of the Association for Computational Linguistics: EMNLP 2024},
  pages={10456--10466},
  year={2024}
}

@article{xie2024moepruner,
  title={Moe-pruner: Pruning mixture-of-experts large language model using the hints from its router},
  author={Xie, Yanyue and Zhang, Zhi and Zhou, Ding and Xie, Cong and Song, Ziang and Liu, Xin and Wang, Yanzhi and Lin, Xue and Xu, An},
  journal={arXiv preprint arXiv:2410.12013},
  year={2024}
}

@article{huang2025mcmoe,
  title={Mixture compressor for mixture-of-experts llms gains more},
  author={Huang, Wei and Liao, Yue and Liu, Jianhui and He, Ruifei and Tan, Haoru and Zhang, Shiming and Li, Hongsheng and Liu, Si and Qi, Xiaojuan},
  journal={arXiv preprint arXiv:2410.06270},
  year={2024}
}

@inproceedings{xu2026seeing,
  title={Seeing but not thinking: Routing distraction in multimodal mixture-of-experts},
  author={Xu, Haolei and Hong, Haiwen and Li, Hongxing and Zhou, Rui and Zhang, Yang and Huang, Longtao and Xue, Hui and Shen, Yongliang and Lu, Weiming and Zhuang, Yueting},
  booktitle={Proceedings of the 64th Annual Meeting of the Association for Computational Linguistics (Volume 1: Long Papers)},
  pages={31164--31178},
  year={2026}
}
